\documentclass{article}
\usepackage[text={170mm,254mm}]{geometry}
\usepackage{graphicx}
\usepackage[T1]{fontenc}
\usepackage[utf8]{inputenc}

\usepackage{amssymb, amsthm, amsfonts, amsmath, enumitem, mathtools}
\usepackage{booktabs, xcolor, multicol}

\usepackage[hyphens]{url}
\usepackage[colorlinks=true,citecolor=black,linkcolor=black,urlcolor=blue]{hyperref}
\usepackage{cite}

\usepackage{subcaption}
\usepackage{float}

\newcommand{\arxiv}[2]{\href{https://arxiv.org/abs/#1}{\texttt{arXiv:#1}} \texttt{[#2]}}
\newcommand{\doi}[1]{\url{https://doi.org/#1}}

\theoremstyle{plain}
\newtheorem{theorem}{Theorem}[section]
\newtheorem{conjecture}[theorem]{Conjecture}
\theoremstyle{definition}
\newtheorem{example}[theorem]{Example}

\usepackage{listings}
\definecolor{mauve}{rgb}{0.58,0,0.82}
\definecolor{dkgreen}{rgb}{0,0.6,0}
\lstdefinestyle{pitonche} {
    language = Python,
    basicstyle = footnotesizettfamily,
    showspaces = false,
    showstringspaces = false,
    breakautoindent = true,
    flexiblecolumns = true,
    keepspaces = true,
    stepnumber = 1,
    xleftmargin = 0pt
}
\let\le=\leqslant
\let\ge=\geqslant

\usepackage{authblk}

\title{RLGT: A reinforcement learning\\ framework for extremal graph theory\thanks{This research was supported and funded by the Ministry of Science, Technological Development and Innovation of the Republic of Serbia, grant number 451-03-137/2025-03/200102, and the Science Fund of the Republic of Serbia, grant \#6767, Lazy walk counts and spectral radius of threshold graphs --- LZWK.}}

\author[1,2]{Ivan Damnjanović\thanks{Corresponding author. Email: \texttt{ivan.damnjanovic@elfak.ni.ac.rs} (I.\ Damnjanović).}}
\author[1]{Uroš Milivojević}
\author[1]{Irena Đorđević}
\author[3]{Dragan Stevanović\thanks{On leave from the Mathematical Institute of the Serbian Academy of Sciences and Arts.}}

\affil[1]{Faculty of Electronic Engineering, University of Niš, Aleksandra Medvedeva 4, Niš, 18104, Serbia}
\affil[2]{Faculty of Mathematics, Natural Sciences and Information Technologies, University of Primorska,\linebreak Glagoljaška 8, Koper, 6000, Slovenia}
\affil[3]{College of Integrative Studies, Abdullah Al Salem University, Firdous Street, Block 3, Khaldiya, 72303, Kuwait}

\date{}

\begin{document}

\maketitle

\begin{abstract}
    Reinforcement learning (RL) is a subfield of machine learning that focuses on developing models that can autonomously learn optimal decision-making strategies over time. In a recent pioneering paper, Wagner demonstrated how the Deep Cross-Entropy RL method can be applied to tackle various problems from extremal graph theory by reformulating them as combinatorial optimization problems. Subsequently, many researchers became interested in refining and extending the framework introduced by Wagner, thereby creating various RL environments specialized for graph theory. Moreover, a number of problems from extremal graph theory were solved through the use of RL. In particular, several inequalities concerning the Laplacian spectral radius of graphs were refuted, new lower bounds were obtained for certain Ramsey numbers, and contributions were made to the Tur\'{a}n-type extremal problem in which the forbidden structures are cycles of length three and four. Here, we present Reinforcement Learning for Graph Theory (RLGT), a novel RL framework that systematizes the previous work and provides support for both undirected and directed graphs, with or without loops, and with an arbitrary number of edge colors. The framework efficiently represents graphs and aims to facilitate future RL-based research in extremal graph theory through optimized computational performance and a clean and modular design.
\end{abstract}

\bigskip\noindent
{\bf Keywords:} reinforcement learning, extremal graph theory, conjecture solving, machine learning.

\bigskip\noindent
{\bf Mathematics Subject Classification:} 68T05, 68T07, 05C35.

\section{Introduction}

Reinforcement learning (RL) is a subfield of machine learning (ML) that deals with developing models that automatically learn optimal decisions over time \cite{Lapan2020}. At a high level, an RL system comprises an agent and an environment, with the agent iteratively interacting with the environment by performing actions on it, and the environment providing feedback in return through observations and rewards. In an RL setting, the agent aims to discover a strategy, called the policy, which should maximize the long-term success of its actions with respect to the rewards returned by the environment. Since the agent learns purely by interacting with the environment without having any additional information on the problem being solved, RL is considered to be much more focused on goal-directed learning through interaction than other ML paradigms \cite{SuBa2018}.

A combinatorial optimization problem is any problem where a given function $f \colon C \to \mathbb{R}$ should be maximized (resp.\ minimized) over a finite set of configurations $C$. As it turns out, the RL formalism can naturally be adapted to tackle such problems; see \cite{MaSviIvBu2021} and the references therein. This can be achieved by considering an RL environment whose states correspond to complete (or partial) configurations, and where the rewards indicate how an action improves or worsens a given configuration with respect to $f$. Here, we consider the applications of RL to solving combinatorial optimization problems pertaining to graphs, i.e., extremal graph theory problems.

Recently, Wagner \cite{Wagner2021} demonstrated how RL can be successfully used to construct counterexamples that refute graph theory conjectures. His idea was to create an RL environment that constructs simple undirected graphs of a given order $n \in \mathbb{N}$ by arranging the $\binom{n}{2}$ unordered pairs of vertices in some manner and executing $\binom{n}{2}$ binary actions that correspond to these pairs. Here, if the $i$-th action is $1$, then the vertices in the $i$-th pair should be adjacent; otherwise, they should not be adjacent. Additionally, a reward is received only after the final action is executed, and it should equal a configurable graph invariant $f$ of the constructed graph. Although such an environment is simple, Wagner showed that the Deep Cross-Entropy method \cite{BoKroMaRu2005, Rubinstein1997} can be used in conjunction with it to achieve satisfactory results for the problem of maximizing a graph invariant $f$ over the set of graphs of a given order. As a direct consequence, it is possible to disprove inequalities involving graphs by transforming the expression $L(G) \le R(G)$ to $L(G) - R(G) \le 0$ and finding a graph whose corresponding invariant is positive, where $L(G)$ (resp.\ $R(G)$) denotes the left-hand (resp.\ right-hand) side of an inequality in graph $G$. With this approach, Wagner disproved several conjectured claims either by directly obtaining a counterexample or by uncovering structural patterns that helped manually construct a counterexample.

Wagner's approach was also successfully used in \cite{SteDaSte2021} to refute a conjecture by Akbari, Alazemi and Anđelić \cite{AkAlaAn2021} on the graph energy and the matching number of graphs. Afterwards, Ghebleh et al.\ \cite{GheYaKaSte2024} offered a reimplementation of Wagner's approach to increase its readability, stability and computational performance. In this framework, the Deep Cross-Entropy method was again used in conjunction with the RL environment introduced by Wagner, but the operations involving states were notably implemented more efficiently through \texttt{NumPy}-based vectorization \cite{NumPy}. Additionally, the final reward function was turned into a separate argument so that it could optionally be executed more efficiently using external code, e.g., \texttt{Java} code using \texttt{JPype} \cite{JPype}. With this approach and by applying the features from the \texttt{graph6java} library \cite{Graph6Java}, the authors succeeded in disproving several previously conjectured upper bounds on the Laplacian spectral radius of graphs \cite{BraHaSte2006}. We briefly note that Taieb et~al.\ \cite{TaRouCaHa2025} successfully refuted two more of these upper bounds by applying the Monte Carlo search technique.

Using the same framework developed in \cite{GheYaKaSte2024}, Ghebleh et al.\ \cite{GheYaKaSte2025} obtained four new lower bounds on small Ramsey numbers involving complete bipartite graphs, wheel graphs and book graphs \cite{Radziszowski}. Afterwards, this framework was used once again by the same authors \cite{GheYaKaSte2026} to help obtain an explicit construction of harmonic graphs \cite{Rowlinson2007} with arbitrarily many distinct vertex degrees.

Concurrently, Mehrabian et al.\ \cite{Mehrabian2024} used RL to tackle the Tur\'{a}n-type extremal problem \cite{Simonovits1984} originally posed by Erdős \cite{Erdos1975} in 1975, in which the forbidden structures are cycles of length three and four. In their approach, a different RL environment was used, where the states are all the graphs of a given order and the actions are edge-flipping operations. By incorporating curriculum learning \cite{SoIoRoSe2022} into the AlphaZero \cite{Silver2017} and tabu search \cite{Glover1989, GloLa1998} algorithms, they obtained new lower bounds for $n \in \{ 64, 65, 66, \ldots, 134 \}$, where $n$ is the graph order. We mention in passing that this was achieved through a novel neural network architecture called the Pairformer.

In a subsequent paper, Angileri et al.\ \cite{Angileri2025} systematized the previous work by implementing four distinct RL environments specialized for graph theory: Linear, Local, Global and Flip. Here, the Linear environment is based on Wagner's original approach \cite{Wagner2021}, while Flip is precisely the edge-flipping environment from the paper of Mehrabian et al.\ \cite{Mehrabian2024}. The four RL environments were implemented in the object-oriented paradigm as classes that inherit from the \texttt{Env} class from the \texttt{Gymnasium} library \cite{Gymnasium}, and their states provide support for finite undirected graphs without multiple edges, with or without loops. Additionally, the authors introduced a novel dataset of graphs labeled with their Laplacian spectra for the purpose of facilitating research involving the Laplacian spectral properties of graphs. Later on, Angileri et al.\ \cite{Angileri2025B} offered a modification of their RL framework and applied it to contribute to the study of Brouwer's conjecture \cite{BrouHae2012}.

We present Reinforcement Learning for Graph Theory (RLGT), a novel RL framework specialized for extremal graph theory that aims to systematize the previous work and bridge the gap between the computationally efficient approach of Ghebleh et al. \cite{GheYaKaSte2024, GheYaKaSte2025, GheYaKaSte2026} and the more expressive and flexible approach of Angileri et al.\ \cite{Angileri2025, Angileri2025B}. The framework is implemented in the object-oriented paradigm as a project in the programming language \texttt{Python}, and it is based on the following principles.
\begin{enumerate}[label=\textbf{(\arabic*)}]
    \item To make the framework fully modularized while keeping the project structure clean, we split the project into three packages: \texttt{graphs}, \texttt{environments} and \texttt{agents}.

    \item The \texttt{graphs} package contains the core class that enables the user to conveniently represent graphs in eight possible formats and automatically perform conversions between these formats. This is an improvement over the previous approaches, where no such class existed and the graph format conversions were left to the end user.

    \item Besides representing graphs, the \texttt{graphs} package is also capable of representing a batch of graphs as a single object. This is inspired by the approach of Ghebleh et al.\ \cite{GheYaKaSte2024, GheYaKaSte2025, GheYaKaSte2026} and it enables the operations involving graphs and states to be performed more efficiently through \texttt{NumPy}-based vectorization.

    \item The graphs and batches of graphs represented in the \texttt{graphs} package can be either undirected or directed, and may or may not have loops. Essentially, the only requirement is that the represented undirected (resp.\ directed) graph is finite and has no multiple edges (resp.\ arcs). Additionally, the edges (resp.\ arcs) can be colored in an arbitrary number of colors, which provides direct support for problems involving edge coloring \cite{CaoChenJingStieToft2019}. This extends the previous work, where there was no support for directed graphs or for graphs with more than two edge colors.
    
    \item The \texttt{environments} package contains the class implementations of RL environments specialized for graph theory. The implemented environments are largely inspired by the approach of Angileri et al.\ \cite{Angileri2025, Angileri2025B}. We provide nine different environments realized as seven classes.
    
    \item The \texttt{agents} package contains the classes corresponding to three different RL methods to be used in conjunction with the available RL environments. This improves the previous frameworks, where either the project structure was not modularized, or the agents were not encapsulated as fully separate entities. The three available RL methods are the Deep Cross-Entropy method \cite{BoKroMaRu2005, Rubinstein1997}, the REINFORCE method \cite{Williams1992}, and the Proximal Policy Optimization (PPO) method \cite{Schulman2017}, and they are all implemented using \texttt{PyTorch} \cite{PyTorch}.

    \item To increase stability and reproducibility, we use the \texttt{Poetry} tool \cite{Poetry} for \texttt{Python} packaging and dependency management. In addition, the code clarity is improved through the \texttt{Black} \cite{Black} and \texttt{isort} \cite{Isort} tools. Finally, the \texttt{pytest} testing framework \cite{Pytest} is applied to unit-test the framework features, additionally increasing the project stability.
\end{enumerate}

In Section \ref{sc_prel}, we present the graph-theoretic and RL foundations on which the proposed framework is based. Afterwards, we give an overview of the developed framework in Section \ref{sc_framework} and provide justifications for many of its implementation details. In Section \ref{sc_applications}, we provide three applications to concrete graph theory problems that demonstrate the framework's ease of use and efficiency. Finally, in Section \ref{sc_conclusion}, we end the paper with a brief conclusion and discuss possible directions for future work. The \texttt{Python} implementation of the presented RL framework can be found in \cite{GitHub}, while the documentation is available in \cite{Documentation} and the Python Package Index (PyPI) page is available in \cite{PyPI}.

\section{Preliminaries}\label{sc_prel}

In this section, we introduce the basic definitions from graph theory and RL that are required to comprehend the proposed framework design and implementation.

\subsection{Graph theory}

All undirected graphs are assumed to be finite and without multiple edges, and we consider all directed graphs to be finite and without multiple arcs. In particular, loops are allowed in both undirected and directed graphs. We use the term \emph{graph} to refer to either an undirected graph or a directed graph, and we denote the vertex set of a graph $G$ by $V(G)$ and the edge set by $E(G)$. The \emph{order} of a graph $G$ is the number of vertices it contains, i.e., $|V(G)|$. For convenience, we assume that $V(G) = \{ 0, 1, 2, \ldots, n - 1 \}$ for any graph $G$ of order $n$. Additionally, we assume that all vector, matrix or tensor indexing is zero-based. The \emph{adjacency matrix} of a graph $G$ of order $n$, denoted by $A(G)$, is the binary matrix in $\mathbb{R}^{n \times n}$ defined by
\[
    A(G)_{u, v} = \begin{cases}
        1, & \mbox{if $u$ is adjacent to $v$},\\
        0, & \mbox{otherwise}
    \end{cases} \qquad (0 \le u, v \le n - 1) .
\]
Recall that if the graph $G$ is undirected, then $A(G)$ is a symmetric matrix. For other undefined terminology from elementary graph theory, the reader can refer to the standard literature \cite{Biggs1993, Bollobas1998, BonMur1976, BrouHae2012, CvetDoobSachs1995, Diestel2017, GodRoy2001}.

For any $k \in \mathbb{N}$, a \emph{$k$-edge-colored looped complete undirected graph} is an undirected graph that contains all possible edges, including loops, with each edge being labeled by a color from the set $\{ 0, 1, 2, \ldots, k \}$. Here, the numbers $0, 1, 2, \ldots, k - 1$ represent the $k$ proper edge colors, while the number $k$ can be optionally used to label an edge that is uncolored, i.e., not colored yet. Similarly, for any $k \in \mathbb{N}$, a \emph{$k$-edge-colored looped complete directed graph} is a directed graph that contains all possible arcs, including loops, with each arc being labeled by a color from the set $\{ 0, 1, 2, \ldots, k \}$. We say that a $k$-edge-colored looped complete undirected (resp.\ directed) graph is \emph{fully colored} if no edge (resp.\ arc) is labeled by the number $k$.

Any undirected graph $G$ can be viewed as a $2$-edge-colored looped complete undirected graph of the same order by coloring the edges of $G$ with the color $1$ and the remaining edges with the color $0$. Therefore, undirected graphs naturally correspond to $2$-edge-colored looped complete undirected graphs. Analogously, directed graphs correspond to $2$-edge-colored looped complete directed graphs. With this in mind, we can use the $k$-edge-colored looped complete graphs to represent both graphs whose edges or arcs are inherently colored and those used in extremal graph theory problems without edge coloring.

\subsection{Reinforcement learning}

In practice, an RL task is typically modeled as a \emph{Markov Decision Process} (MDP), defined as an ordered triple $(\mathcal{S}, \mathcal{A}, p)$, where $\mathcal{S}$ is the state space, $\mathcal{A}$ is the action space, and $p \colon \mathcal{S} \times \mathcal{A} \to \Delta (\mathcal{S} \times \mathbb{R})$ is the transition function, with $\Delta(\cdot)$ denoting the set of probability distributions over its argument. Here, $p(s', r \mid s, a)$ is the probability of reaching the next state $s'$ and receiving the reward $r$ after the action $a$ is executed in the current state $s$. A \emph{policy} is a strategy used by the agent to select actions while interacting with the environment, formalized as a function $\pi \colon \mathcal{S} \to \Delta(\mathcal{A})$. An \emph{episodic task} is an RL task where a terminal state is eventually reached regardless of how the agent acts on the environment, whereas a \emph{continuing task} is one without terminal states, in which the agent--environment interaction can proceed indefinitely. For further details on RL theory and terminology, the reader can refer to the standard textbooks \cite{Lapan2020, Powell2022, SuBa2018, Szepesvari2010}.

We consider the problem of maximizing the function $f$ over the set of fully colored $k$-edge-colored looped complete (undirected or directed) graphs of order $n$, for a given $n \ge 2$, $k \ge 2$ and graph invariant $f$. We are thus only interested in RL tasks designed to build such graphs. To begin, we consider only deterministic tasks, which means that whenever an action $a$ is executed in a state $s$, the same next state $s'$ is always reached and the same reward $r$ is always received. Therefore, we can formalize the transition model through a function of the form $\mathcal{S} \times \mathcal{A} \to \mathcal{S} \times \mathbb{R}$ instead of the more general $\mathcal{S} \times \mathcal{A} \to \Delta (\mathcal{S} \times \mathbb{R})$. Besides, we assume that an RL task is either continuing, like the edge-flipping environment introduced by Mehrabian et al.\ \cite{Mehrabian2024}, or it is episodic and a terminal state is reached after a predetermined number of actions, regardless of how the actions are selected, like the environment from Wagner's original approach \cite{Wagner2021}. In the case of continuing tasks, the RL environment receives a parameter that determines the total number of actions to be executed within each episode.

To make the graph invariant $f$ a fully configurable parameter of the RL environment, we separate the logic behind the state transition and the reward computation. For any choice of $f$, we assume that the same state $s'$ is reached from a state $s$ when an action $a$ is executed. Hence, the state transition can be modeled through a function of the form $t \colon \mathcal{S} \times \mathcal{A} \to \mathcal{S}$, where $s' = t(s, a)$. Additionally, we assume that each state $s$ has an underlying (not necessarily fully colored) $k$-edge-colored looped complete graph $\varphi(s)$.

As for the reward computation, our approach is to use graph invariant values rather than conventional RL rewards. In other words, when the environment reaches a state $s$, the agent receives the value $f(\varphi(s))$. We believe that such an agent--environment interaction is natural in the context of tackling extremal graph theory problems. In addition, we recognize two types of agent--environment communication settings: sparse and dense. If an RL environment uses the \emph{sparse setting}, then the attained graph invariant is received only after the final action, and $f(\varphi(s))$ need not be defined for non-final states $s$. Conversely, if the environment uses the \emph{dense setting}, then the attained graph invariant is received after each executed action. In the previous work, Wagner \cite{Wagner2021} and Ghebleh et al.\ \cite{GheYaKaSte2024, GheYaKaSte2025, GheYaKaSte2026} used the sparse setting, while Mehrabian et al.\ \cite{Mehrabian2024} essentially used the dense setting through telescopic rewards. Our approach is partially inspired by Angileri et al.\ \cite{Angileri2025, Angileri2025B}, hence we support both communication settings in a modularized and clean manner.

\section{Framework overview}\label{sc_framework}

The presented RL framework was implemented in the programming language \texttt{Python}. Although there are more computationally efficient languages, such as \texttt{C}, \texttt{C++}, \texttt{C\#}, \texttt{Java}, \texttt{Rust} and \texttt{Go} \cite{NanzFuria2015}, our choice of language is justified by the full support of the two well-known deep learning libraries \texttt{PyTorch} \cite{PyTorch} and \texttt{TensorFlow} \cite{TensorFlow} in \texttt{Python}. Additionally, the vectorized operations in \texttt{NumPy} provide satisfactory computational efficiency, especially when executed on a modern high-end computer. Finally, due to the expressive power of \texttt{Python}, the framework can be used without extensive prior programming experience, making it accessible to a wide spectrum of end users. The framework implementation can be found in \cite{GitHub}, while the documentation is available in \cite{Documentation} and the PyPI page is available in \cite{PyPI}.

For the sake of modularity, the framework is split into three packages: \texttt{graphs}, \texttt{environments} and \texttt{agents}. The \texttt{graphs} package encapsulates graphs and batches of graphs, representing them in eight possible formats and automatically performing all required conversions between these formats. The \texttt{environments} package contains nine RL environments implemented as seven classes that provide support for various graph-building games. Additionally, this package contains several auxiliary functions that help create deterministic or nondeterministic graph generators. The \texttt{agents} package contains the three RL agents corresponding to the Deep Cross-Entropy, REINFORCE, and PPO methods, alongside several classes that implement random action mechanisms. The dependencies between the three packages are clean, with the \texttt{graphs} package having no dependencies on the other two packages, and the \texttt{environments} package depending only on the \texttt{graphs} package. Therefore, the packages follow a layered design.

The \texttt{Poetry} tool is used for \texttt{Python} packaging and dependency management. The first two packages, \texttt{graphs} and \texttt{environments}, have no external dependencies apart from \texttt{NumPy}, while the third package, \texttt{agents}, additionally depends only on \texttt{PyTorch}. For extendability, only \texttt{NumPy} is considered an obligatory dependency, while \texttt{PyTorch} is not installed by default. This allows the end user to potentially employ another deep learning library instead of \texttt{PyTorch}. For instance, if the user prefers \texttt{TensorFlow} to \texttt{PyTorch}, they can perform the default installation and use only the \texttt{graphs} and \texttt{environments} packages together with their own RL methods implemented in \texttt{TensorFlow}. Alternatively, \texttt{PyTorch} can be installed as an optional dependency, enabling the use of the three available RL methods, all of which are implemented in \texttt{PyTorch}. This approach highlights the modularity of the framework and is natural, since having two deep learning libraries installed at the same time is typically undesirable, especially if only one of them is used. To increase the project stability, \texttt{pytest} is applied to unit-test the framework features. We additionally use \texttt{Black} and \texttt{isort} to improve the code clarity.

\subsection{Graphs and graph formats}

The core component of the \texttt{graphs} package is the \texttt{Graph} class, which encapsulates the concept of a $k$-edge-colored looped complete graph. The class essentially behaves as a wrapper around a collection of eight \texttt{NumPy} arrays, each of which represents the graph in one of the eight supported graph formats. Apart from several properties, the main functionality that distinguishes the class from a pure octuple is the automatic conversion between these eight formats. We view a $k$-edge-colored looped complete graph as a quintuple \texttt{(edge\_colors, is\_directed, allow\_loops, graph\_format, format\_representation)}, where:
\begin{enumerate}[label=\textbf{(\arabic*)}]
    \item \texttt{edge\_colors} is the number of proper edge colors, i.e., $k$, with the requirement that $k \ge 2$;
    \item \texttt{is\_directed} is a boolean indicating whether the considered graph is a $k$-edge-colored looped complete directed graph or a $k$-edge-colored looped complete undirected graph;
    \item \texttt{allow\_loops} is a boolean indicating whether the considered graph is allowed to have loops;
    \item \texttt{graph\_format} is one of the eight supported graph formats; and
    \item \texttt{format\_representation} is the \texttt{NumPy} array representing the structure of the considered graph in the chosen graph format.
\end{enumerate}

Although the presence of loops can be inferred from the graph structure, we use this representation because the elements \texttt{edge\_colors}, \texttt{is\_directed} and \texttt{allow\_loops} directly affect how the \texttt{format\_representation} \texttt{NumPy} array is used to obtain the considered graph. We provide support for the following eight graph formats.
\begin{enumerate}[label=\textbf{(\arabic*)}]
    \item The \emph{bitmask format for the out-neighborhoods} represents the graph structure through a matrix $B \in \mathbb{Z}^{k \times n}$, where $k$ is the number of proper edge colors and $n$ is the graph order. All entries of $B$ are integers between $0$ and $2^n - 1$, so that for any $c \in \{0, 1, 2, \ldots, k - 1 \}$ and $u, v \in \{ 0, 1, 2, \ldots, n - 1 \}$, the $v$-th bit in the binary representation of $B_{c, u}$ is one if and only if the edge (resp.\ arc) from vertex $u$ to vertex $v$ has color $c$. If loops are not allowed, then the $u$-th bit of $B_{c, u}$ is taken to be zero for any $c \in \{ 0, 1, 2, \ldots, k - 1\}$ and $u \in \{ 0, 1, 2, \ldots, n - 1 \}$.
    
    \item The \emph{bitmask format for the in-neighborhoods} represents the graph structure in the same way as the bitmask format for the out-neighborhoods, with the difference that the $v$-th bit of $B_{c, u}$ indicates whether the edge (resp.\ arc) from vertex $v$ to vertex $u$ has color $c$, instead of the edge (resp.\ arc) from vertex $u$ to vertex $v$. These two formats coincide if the considered graph is undirected.

    \item The \emph{adjacency matrix format with color numbers} employs a variant of the adjacency matrix to represent the graph structure. More precisely, this format uses a matrix $A \in \mathbb{Z}^{n \times n}$, where $n$ is the graph order, such that for any $u, v \in \{ 0, 1, 2, \ldots, n - 1 \}$, the entry $A_{u, v}$ is equal to the color of the edge (resp.\ arc) from vertex $u$ to vertex $v$. Recall that an uncolored edge (resp.\ arc) is represented by the color $k$, where $k$ is the number of proper edge colors. If loops are not allowed, then the diagonal entries of $A$ are all equal to zero. Moreover, if the considered graph is undirected, then the matrix $A$ is symmetric.

    \item The \emph{adjacency matrix format with binary slices} is similar to the adjacency matrix format with color numbers, with the difference that the colors correspond to separate binary matrices instead of being represented by integer values in a single matrix. More precisely, this format employs a binary tensor $\mathbf{A} \in \mathbb{Z}^{k \times n \times n}$, where $k$ is the number of proper edge colors and $n$ is the graph order, such that for any $c \in \{ 0, 1, 2, \ldots, k - 1 \}$ and $u, v \in \{ 0, 1, 2, \ldots, n - 1 \}$, the entry $\mathbf{A}_{c, u, v}$ equals one if and only if the edge (resp.\ arc) from vertex $u$ to vertex $v$ has color $c$. If loops are not allowed, then $\mathbf{A}_{c, u, u} = 0$ for any $c \in \{ 0, 1, 2, \ldots, k - 1 \}$ and $u \in \{ 0, 1, 2, \ldots, n - 1 \}$.

    \item The \emph{flattened row-major format with color numbers} represents the graph structure through a vector in $\mathbb{Z}^\ell$ containing the entries of the matrix $A$ from the adjacency matrix format with color numbers, arranged in row-major order. In other words, the entries are arranged so that the first row is traversed from left to right, then the second row is traversed from left to right, and so on until the last row. Additionally, all redundancy in data storage is avoided. More precisely, if the considered graph is directed and loops are not allowed, then the diagonal entries are omitted. If the graph is undirected, then only the entries from the upper triangular part of $A$ are arranged in row-major order, with or without the diagonal, depending on whether loops are allowed. Therefore,
    \begin{equation}\label{flattened_length_eq}
        \ell = \begin{cases}
            n^2 , & \mbox{if the considered graph is directed and loops are allowed},\\
            n^2 - n, & \mbox{if the considered graph is directed and loops are not allowed},\\
            \binom{n + 1}{2}, & \mbox{if the considered graph is undirected and loops are allowed},\\
            \binom{n}{2}, & \mbox{if the considered graph is undirected and loops are not allowed} ,
        \end{cases}
    \end{equation}
    where $n$ is the graph order.

    \item The \emph{flattened row-major format with binary slices} is similar to the flattened row-major format with color numbers, with the difference that the colors correspond to separate binary vectors instead of being represented by integer values in a single vector. This format employs a binary matrix $F \in \mathbb{Z}^{k \times \ell}$, where $k$ is the number of proper edge colors and $\ell$ is given by \eqref{flattened_length_eq}, such that for any $c \in \{ 0, 1, 2, \ldots, k - 1\}$ and $i \in \{ 0, 1, 2, \ldots, \ell - 1 \}$, the entry $F_{c, i}$ equals one if and only if the $i$-th edge (resp.\ arc) has color $c$. Here, the edges (resp.\ arcs) are assumed to be arranged in row-major order.

    \item The \emph{flattened clockwise format with color numbers} represents the graph structure in the same way as the flattened row-major format with color numbers, with the difference that the edges (resp.\ arcs) are arranged in clockwise order instead of row-major order. By clockwise order, we mean the order
    \[
        (0, 0), (0, 1), (1, 1), (1, 0), (0, 2), (1, 2), (2, 2), (2, 1), (2, 0), \ldots,
    \]
    where the entries are traversed in a clockwise layer-like manner. Note that if the considered graph is undirected, then the obtained order can equivalently be regarded as the column-major order over the upper triangular part of the adjacency matrix.

    \item The \emph{flattened clockwise format with binary slices} uses the same approach to represent the graph structure as the flattened row-major format with binary slices, with the difference that the edges (resp.\ arcs) are arranged in clockwise order instead of row-major order.
\end{enumerate}

To further reduce redundancy in data storage, we can omit the entries corresponding to color zero in the bitmask formats and the formats with binary slices, provided the considered graph is fully colored. For example, a bitmask format representation of a fully colored graph can be given as a matrix in $\mathbb{Z}^{(k - 1) \times n}$ instead of $\mathbb{Z}^{k \times n}$, where the rows correspond to colors $1, 2, \ldots, k - 1$, respectively. We refer to such a format as a \emph{reduced format}. These format variants are well defined because if the considered graph is fully colored, then an edge has color zero if it has no color from $\{1, 2, \ldots, k - 1\}$, and we can infer whether the given format is reduced from the shape of the \texttt{format\_representation} \texttt{NumPy} array. Moreover, it is natural to ignore color zero for practical reasons in the frequent case where $k = 2$. Two examples of $k$-edge-colored looped complete directed graphs in all eight supported formats are shown in Table \ref{directed_table}, and two examples of $k$-edge-colored looped complete undirected graphs in all eight supported formats are shown in Table \ref{undirected_table}.

\begin{table}[t]
\centering
\begin{tabular}{|l|c|c|c|c|}
\hline
& Graph $G_1$ & Graph $G_2$ \\
\hline
\texttt{edge\_colors} & 3 & 3\\
\hline
\texttt{is\_directed} & \texttt{True} & \texttt{True} \\
\hline
\texttt{allow\_loops} & \texttt{True} & \texttt{False} \\
\hline
\begin{tabular}{@{}l@{}}
Bitmask format\\for the out-neighborhoods
\end{tabular} & $\begin{bmatrix}\begin{smallmatrix} 8 & 8 & 0 & 0\\ 0 & 0 & 4 & 9\\ 7 & 0 & 10 & 2 \end{smallmatrix}\end{bmatrix}$ & $\begin{bmatrix}\begin{smallmatrix} 8 & 12 & 0 & 0\\ 6 & 0 & 9 & 2 \end{smallmatrix}\end{bmatrix}$ \\
\hline
\begin{tabular}{@{}l@{}}
Bitmask format\\for the in-neighborhoods
\end{tabular} & $\begin{bmatrix}\begin{smallmatrix} 0 & 0 & 0 & 3\\ 8 & 0 & 4 & 8\\ 1 & 13 & 1 & 4 \end{smallmatrix}\end{bmatrix}$ & $\begin{bmatrix}\begin{smallmatrix} 0 & 0 & 2 & 3\\ 4 & 9 & 1 & 4 \end{smallmatrix}\end{bmatrix}$ \\
\hline
\begin{tabular}{@{}l@{}}
Adjacency matrix format\\with color numbers
\end{tabular} & $\rule{0pt}{4ex} \begin{bmatrix}\begin{smallmatrix} 2 & 2 & 2 & 0\\ 3 & 3 & 3 & 0\\ 3 & 2 & 1 & 2\\ 1 & 2 & 3 & 1 \end{smallmatrix}\end{bmatrix} \rule[-2.8ex]{0pt}{0ex}$ & $\rule{0pt}{4ex} \begin{bmatrix}\begin{smallmatrix} 0 & 2 & 2 & 1\\ 0 & 0 & 1 & 1\\ 2 & 0 & 0 & 2\\ 0 & 2 & 0 & 0 \end{smallmatrix}\end{bmatrix} \rule[-2.8ex]{0pt}{0ex}$ \\
\hline
\begin{tabular}{@{}l@{}}
Adjacency matrix format\\with binary slices
\end{tabular} & $\rule{0pt}{4ex} \left[ \begin{bmatrix}\begin{smallmatrix} 0 & 0 & 0 & 1\\ 0 & 0 & 0 & 1\\ 0 & 0 & 0 & 0\\ 0 & 0 & 0 & 0 \end{smallmatrix}\end{bmatrix}, \begin{bmatrix}\begin{smallmatrix} 0 & 0 & 0 & 0\\ 0 & 0 & 0 & 0\\ 0 & 0 & 1 & 0\\ 1 & 0 & 0 & 1 \end{smallmatrix}\end{bmatrix}, \begin{bmatrix}\begin{smallmatrix} 1 & 1 & 1 & 0\\ 0 & 0 & 0 & 0\\ 0 & 1 & 0 & 1\\ 0 & 1 & 0 & 0 \end{smallmatrix}\end{bmatrix}\right] \rule[-2.8ex]{0pt}{0ex}$ & $\left[ \begin{bmatrix}\begin{smallmatrix} 0 & 0 & 0 & 1\\ 0 & 0 & 1 & 1\\ 0 & 0 & 0 & 0\\ 0 & 0 & 0 & 0 \end{smallmatrix}\end{bmatrix}, \begin{bmatrix}\begin{smallmatrix} 0 & 1 & 1 & 0\\ 0 & 0 & 0 & 0\\ 1 & 0 & 0 & 1\\ 0 & 1 & 0 & 0 \end{smallmatrix}\end{bmatrix} \right]$ \\
\hline
\begin{tabular}{@{}l@{}}
Flattened row-major format\\with color numbers
\end{tabular} & $\begin{bmatrix}\begin{smallmatrix} 2 & 2 & 2 & 0 & 3 & 3 & 3 & 0 & 3 & 2 & 1 & 2 & 1 & 2 & 3 & 1 \end{smallmatrix}\end{bmatrix}^\intercal$ & $\begin{bmatrix}\begin{smallmatrix} 2 & 2 & 1 & 0 & 1 & 1 & 2 & 0 & 2 & 0 & 2 & 0 \end{smallmatrix}\end{bmatrix}^\intercal$ \\
\hline
\begin{tabular}{@{}l@{}}
Flattened row-major format\\with binary slices
\end{tabular} & $\begin{bmatrix}\begin{smallmatrix} 0 & 0 & 0 & 1 & 0 & 0 & 0 & 1 & 0 & 0 & 0 & 0 & 0 & 0 & 0 & 0\\ 0 & 0 & 0 & 0 & 0 & 0 & 0 & 0 & 0 & 0 & 1 & 0 & 1 & 0 & 0 & 1\\ 1 & 1 & 1 & 0 & 0 & 0 & 0 & 0 & 0 & 1 & 0 & 1 & 0 & 1 & 0 & 0 \end{smallmatrix}\end{bmatrix}$ & $\begin{bmatrix}\begin{smallmatrix} 0 & 0 & 1 & 0 & 1 & 1 & 0 & 0 & 0 & 0 & 0 & 0\\ 1 & 1 & 0 & 0 & 0 & 0 & 1 & 0 & 1 & 0 & 1 & 0 \end{smallmatrix}\end{bmatrix}$ \\
\hline
\begin{tabular}{@{}l@{}}
Flattened clockwise format\\with color numbers
\end{tabular} & $\begin{bmatrix}\begin{smallmatrix} 2 & 2 & 3 & 3 & 2 & 3 & 1 & 2 & 3 & 0 & 0 & 2 & 1 & 3 & 2 & 1 \end{smallmatrix}\end{bmatrix}^\intercal$ & $\begin{bmatrix}\begin{smallmatrix} 2 & 0 & 2 & 1 & 0 & 2 & 1 & 1 & 2 & 0 & 2 & 0 \end{smallmatrix}\end{bmatrix}^\intercal$  \\
\hline
\begin{tabular}{@{}l@{}}
Flattened clockwise format\\with binary slices
\end{tabular} & $\begin{bmatrix}\begin{smallmatrix} 0 & 0 & 0 & 0 & 0 & 0 & 0 & 0 & 0 & 1 & 1 & 0 & 0 & 0 & 0 & 0\\ 0 & 0 & 0 & 0 & 0 & 0 & 1 & 0 & 0 & 0 & 0 & 0 & 1 & 0 & 0 & 1\\ 1 & 1 & 0 & 0 & 1 & 0 & 0 & 1 & 0 & 0 & 0 & 1 & 0 & 0 & 1 & 0 \end{smallmatrix}\end{bmatrix}$ & $\begin{bmatrix}\begin{smallmatrix} 0 & 0 & 0 & 1 & 0 & 0 & 1 & 1 & 0 & 0 & 0 & 0\\ 1 & 0 & 1 & 0 & 0 & 1 & 0 & 0 & 1 & 0 & 1 & 0 \end{smallmatrix}\end{bmatrix}$\\
\hline
\end{tabular}
\caption{A directed graph $G_1$ with allowed loops and a directed graph $G_2$ without allowed loops, both represented in all eight supported graph formats. A reduced format is always used when possible.}
\label{directed_table}
\end{table}

\begin{table}[ht]
\centering
\begin{tabular}{|l|c|c|c|c|}
\hline
& Graph $G_3$ & Graph $G_4$ \\
\hline
\texttt{edge\_colors} & 4 & 2\\
\hline
\texttt{is\_directed} & \texttt{False} & \texttt{False} \\
\hline
\texttt{allow\_loops} & \texttt{True} & \texttt{False} \\
\hline
\begin{tabular}{@{}l@{}}
Bitmask format\\for the out-neighborhoods
\end{tabular} & $\rule{0pt}{4ex} \begin{bmatrix}\begin{smallmatrix} 2 & 1 & 0\\ 1 & 4 & 2\\ 4 & 0 & 1\\ 0 & 2 & 0 \end{smallmatrix}\end{bmatrix} \rule[-2.8ex]{0pt}{0ex}$ & $\begin{bmatrix}\begin{smallmatrix} 12 & 24 & 25 & 23 & 14 \end{smallmatrix}\end{bmatrix}$ \\
\hline
\begin{tabular}{@{}l@{}}
Bitmask format\\for the in-neighborhoods
\end{tabular} & $\rule{0pt}{4ex} \begin{bmatrix}\begin{smallmatrix} 2 & 1 & 0\\ 1 & 4 & 2\\ 4 & 0 & 1\\ 0 & 2 & 0 \end{smallmatrix}\end{bmatrix} \rule[-2.8ex]{0pt}{0ex}$ & $\begin{bmatrix}\begin{smallmatrix} 12 & 24 & 25 & 23 & 14 \end{smallmatrix}\end{bmatrix}$ \\
\hline
\begin{tabular}{@{}l@{}}
Adjacency matrix format\\with color numbers
\end{tabular} & $\begin{bmatrix}\begin{smallmatrix} 1 & 0 & 2\\ 0 & 3 & 1\\ 2 & 1 & 4 \end{smallmatrix}\end{bmatrix}$ & $\rule{0pt}{4.6ex} \begin{bmatrix}\begin{smallmatrix} 0 & 0 & 1 & 1 & 0\\ 0 & 0 & 0 & 1 & 1\\ 1 & 0 & 0 & 1 & 1\\ 1 & 1 & 1 & 0 & 1\\ 0 & 1 & 1 & 1 & 0 \end{smallmatrix}\end{bmatrix} \rule[-3.4ex]{0pt}{0ex}$ \\
\hline
\begin{tabular}{@{}l@{}}
Adjacency matrix format\\with binary slices
\end{tabular} & $\left[ \begin{bmatrix}\begin{smallmatrix} 0 & 1 & 0\\ 1 & 0 & 0\\ 0 & 0 & 0 \end{smallmatrix}\end{bmatrix}, \begin{bmatrix}\begin{smallmatrix} 1 & 0 & 0\\ 0 & 0 & 1\\ 0 & 1 & 0 \end{smallmatrix}\end{bmatrix}, \begin{bmatrix}\begin{smallmatrix} 0 & 0 & 1\\ 0 & 0 & 0\\ 1 & 0 & 0 \end{smallmatrix}\end{bmatrix}, \begin{bmatrix}\begin{smallmatrix} 0 & 0 & 0\\ 0 & 1 & 0\\ 0 & 0 & 0 \end{smallmatrix}\end{bmatrix} \right]$ & $\rule{0pt}{4.6ex} \left[ \begin{bmatrix}\begin{smallmatrix} 0 & 0 & 1 & 1 & 0\\ 0 & 0 & 0 & 1 & 1\\ 1 & 0 & 0 & 1 & 1\\ 1 & 1 & 1 & 0 & 1\\ 0 & 1 & 1 & 1 & 0 \end{smallmatrix}\end{bmatrix} \right] \rule[-3.4ex]{0pt}{0ex}$ \\
\hline
\begin{tabular}{@{}l@{}}
Flattened row-major format\\with color numbers
\end{tabular} & $\begin{bmatrix}\begin{smallmatrix} 1 & 0 & 2 & 3 & 1 & 4 \end{smallmatrix}\end{bmatrix}^\intercal$ & $\begin{bmatrix}\begin{smallmatrix} 0 & 1 & 1 & 0 & 0 & 1 & 1 & 1 & 1 & 1 \end{smallmatrix}\end{bmatrix}^\intercal$ \\
\hline
\begin{tabular}{@{}l@{}}
Flattened row-major format\\with binary slices
\end{tabular} & $\rule{0pt}{4ex} \begin{bmatrix}\begin{smallmatrix} 0 & 1 & 0 & 0 & 0 & 0\\ 1 & 0 & 0 & 0 & 1 & 0\\ 0 & 0 & 1 & 0 & 0 & 0\\ 0 & 0 & 0 & 1 & 0 & 0 \end{smallmatrix}\end{bmatrix} \rule[-2.8ex]{0pt}{0ex}$ & $\begin{bmatrix}\begin{smallmatrix} 0 & 1 & 1 & 0 & 0 & 1 & 1 & 1 & 1 & 1 \end{smallmatrix}\end{bmatrix}$ \\
\hline
\begin{tabular}{@{}l@{}}
Flattened clockwise format\\with color numbers
\end{tabular} & $\begin{bmatrix}\begin{smallmatrix} 1 & 0 & 3 & 2 & 1 & 4 \end{smallmatrix}\end{bmatrix}^\intercal$ & $\begin{bmatrix}\begin{smallmatrix} 0 & 1 & 0 & 1 & 1 & 1 & 0 & 1 & 1 & 1 \end{smallmatrix}\end{bmatrix}^\intercal$  \\
\hline
\begin{tabular}{@{}l@{}}
Flattened clockwise format\\with binary slices
\end{tabular} & $\rule{0pt}{4ex} \begin{bmatrix}\begin{smallmatrix} 0 & 1 & 0 & 0 & 0 & 0\\ 1 & 0 & 0 & 0 & 1 & 0\\ 0 & 0 & 0 & 1 & 0 & 0\\ 0 & 0 & 1 & 0 & 0 & 0 \end{smallmatrix}\end{bmatrix} \rule[-2.8ex]{0pt}{0ex}$ & $\begin{bmatrix}\begin{smallmatrix} 0 & 1 & 0 & 1 & 1 & 1 & 0 & 1 & 1 & 1 \end{smallmatrix}\end{bmatrix}$\\
\hline
\end{tabular}
\caption{An undirected graph $G_3$ with allowed loops and an undirected graph $G_4$ without allowed loops, both represented in all eight supported graph formats. A reduced format is always used when possible.}
\label{undirected_table}
\end{table}

Instances of the \texttt{Graph} class are initialized through a provided representation quintuple \texttt{(edge\_colors, is\_directed, allow\_loops, graph\_format, format\_representation)}. Additionally, the end user can initialize an instance in more than one graph format, in which case all provided format representations need to be consistent with one another, i.e., they must represent the same graph. Afterwards, the graph can be accessed in any of the eight supported formats by simply using the corresponding property, with all format conversions being performed automatically. When converting to a bitmask format or a format with binary slices, the reduced format variant is always used when possible.

\begin{example}\label{graph_example}
The graph $G_1$ from Table \ref{directed_table} can be initialized using the \texttt{Graph} constructor as follows.
\begin{lstlisting}[language = Python, frame = trBL, escapeinside={(*@}{@*)}, aboveskip=10pt, belowskip=10pt, numbers=left, rulecolor=\color{black}]
flattened_row_major_colors = np.array(
    [2, 2, 2, 0, 3, 3, 3, 0, 3, 2, 1, 2, 1, 2, 3, 1],
    dtype=np.uint8,
)
g1 = Graph(
    edge_colors=3,
    is_directed=True,
    allow_loops=True,
    flattened_row_major_colors=flattened_row_major_colors,
)
\end{lstlisting}
Here, the graph is initialized in the flattened row-major format with color numbers, and it can then be accessed in any of the eight supported graph formats.
\begin{lstlisting}[language = Python, frame = trBL, escapeinside={(*@}{@*)}, aboveskip=10pt, belowskip=10pt, numbers=left, rulecolor=\color{black}]
print(g1.bitmask_out)
print(g1.bitmask_in)
print(g1.adjacency_matrix_colors)
print(g1.adjacency_matrix_binary)
print(g1.flattened_row_major_colors)
print(g1.flattened_row_major_binary)
print(g1.flattened_clockwise_colors)
print(g1.flattened_clockwise_binary)
\end{lstlisting}
In addition, the \texttt{Graph} class contains three class methods that enable the user to instantiate a \texttt{Graph} object in exactly one specific type of graph format. For example, the graph $G_2$ from Table \ref{directed_table} can be initialized in the bitmask format for the out-neighborhoods as follows.
\begin{lstlisting}[language = Python, frame = trBL, escapeinside={(*@}{@*)}, aboveskip=10pt, belowskip=10pt, numbers=left, rulecolor=\color{black}]
bitmask = np.array(
    [
        [8, 12, 0, 0],
        [6, 0, 9, 2],
    ],
    dtype=np.uint64,
)
g2 = Graph.from_bitmask(
    bitmask=bitmask,
    bitmask_type=BitmaskType.OUT_NEIGHBORS,
    edge_colors=3,
    is_directed=True,
    allow_loops=False,
)
\end{lstlisting}
The following code snippet initializes the graph $G_3$ from Table \ref{undirected_table} in the adjacency matrix format with binary slices.
\pagebreak
\begin{lstlisting}[language = Python, frame = trBL, escapeinside={(*@}{@*)}, aboveskip=10pt, belowskip=10pt, numbers=left, rulecolor=\color{black}]
adjacency_matrix = np.array(
    [
        [
            [0, 1, 0],
            [1, 0, 0],
            [0, 0, 0],
        ],
        [
            [1, 0, 0],
            [0, 0, 1],
            [0, 1, 0],
        ],
        [
            [0, 0, 1],
            [0, 0, 0],
            [1, 0, 0],
        ],
        [
            [0, 0, 0],
            [0, 1, 0],
            [0, 0, 0],
        ],
    ],
    dtype=np.uint8,
)
g3 = Graph.from_adjacency_matrix(
    adjacency_matrix=adjacency_matrix,
    color_representation=ColorRepresentation.BINARY_SLICES,
    edge_colors=4,
    is_directed=False,
    allow_loops=True,
)
\end{lstlisting}
The graph $G_4$ from Table \ref{undirected_table} can be initialized in the flattened clockwise format with color numbers as follows.
\begin{lstlisting}[language = Python, frame = trBL, escapeinside={(*@}{@*)}, aboveskip=10pt, belowskip=10pt, numbers=left, rulecolor=\color{black}]
flattened = np.array([0, 1, 0, 1, 1, 1, 0, 1, 1, 1], dtype=np.uint8)
g4 = Graph.from_flattened(
    flattened=flattened,
    flattened_ordering=FlattenedOrdering.CLOCKWISE,
    color_representation=ColorRepresentation.COLOR_NUMBERS,
)
\end{lstlisting}
Note that by default, a graph is considered undirected, without loops, and with two proper edge colors. Therefore, all the corresponding parameters can be omitted when the default values are sufficient. The full source code for Example \ref{graph_example} is available in the file \texttt{examples/graph\_examples.py} in \cite{GitHub}. ~\hfill $\Diamond$
\end{example}

Besides $k$-edge-colored looped complete graphs, the \texttt{Graph} class also encapsulates batches of $k$-edge-colored looped complete graphs. In this case, the graphs in the batch are required to be of the same order and with the same number of proper edge colors, and they need to be of the same type: they are either all directed or all undirected, and either all of them have allowed loops or none of them have allowed loops. This enables such batches of graphs to be representable using the same eight graph formats by simply using a \texttt{NumPy} array of one dimension higher. In such format representations, the leading dimension corresponds to the graphs in the batch, while the remaining dimensions are used in the same way as when representing single graphs. It is easy to infer whether a \texttt{Graph} object represents a graph or a batch of graphs through the \texttt{batch\_size} property --- for batches of graphs, it returns the number of graphs in the batch, while for single graphs, it returns \texttt{None}.

In addition to the \texttt{Graph} class, the \texttt{graphs} package also contains several classes that inherit from this class and encapsulate $k$-edge-colored looped complete graphs with some particular structure. We provide the following classes that inherit from the \texttt{Graph} class:
\begin{multicols}{3}
\begin{enumerate}[label=\textbf{(\arabic*)}]
    \item \texttt{MonochromaticGraph};
    \item \texttt{EmptyGraph};
    \item \texttt{CompleteGraph};
    \item \texttt{AlmostCompleteGraph};
    \item \texttt{CompleteBipartiteGraph};
    \item \texttt{CompleteKPartiteGraph};
    \item \texttt{StarGraph};
    \item \texttt{PathGraph};
    \item \texttt{CycleGraph};
    \item \texttt{WheelGraph};
    \item \texttt{BookGraph}; and
    \item \texttt{FriendshipGraph}.
\end{enumerate}
\end{multicols}
The name of each of these classes determines the kind of graphs it encapsulates. We mention in passing that all of these graphs have two proper edge colors, besides the graphs encapsulated by the \texttt{MonochromaticGraph} class, which can have any number of proper edge colors. For more details, the reader can refer to the framework documentation \cite{Documentation}.

\subsection{RL environments}

The \texttt{environments} package contains the abstract class \texttt{GraphEnvironment}, which encapsulates RL environments in extremal graph theory applications, alongside several concrete classes that inherit from this class. The two primary methods of the \texttt{GraphEnvironment} class are \texttt{reset\_batch}, which initializes a batch of episodes with a given batch size, and \texttt{step\_batch}, which takes a batch of actions and applies them element-wise to the current states from the ongoing episodes. These two methods follow a \texttt{Gymnasium}-like API and naming convention, with the exception that they operate in batch mode. In other words, multiple episodes can be run and acted on in parallel to increase efficiency through \texttt{NumPy}-based vectorization. This strategy is inspired by the approach of Ghebleh et al.\ \cite{GheYaKaSte2024, GheYaKaSte2025, GheYaKaSte2026}, and it relies on our assumption that all episodes end after a predetermined number of actions, regardless of whether the environment has continuing or episodic RL tasks.

The states in an RL environment are represented as \texttt{NumPy} vectors of some length and type specific to the concrete class that inherits from \texttt{GraphEnvironment}. Naturally, batches of states are then represented as \texttt{NumPy} matrices of the required shape and type, with rows corresponding to the states in the batch. It is assumed that there are finitely many actions, and they are represented by nonnegative integers from $\{ 0, 1, 2, \ldots, q - 1 \}$, where $q \ge 2$ is the total number of actions. Similarly, batches of actions are represented as \texttt{NumPy} vectors of type \texttt{numpy.int32} whose entries correspond to the actions. We mention in passing that not every action needs to be available for execution at every step, but at least one action should be available in any non-terminal state.

Although the \texttt{reset\_batch} and \texttt{step\_batch} methods serve entirely different purposes, they both return the same triple:
\begin{enumerate}[label=\textbf{(\arabic*)}]
    \item the batch of states, obtained upon initialization or after applying the provided actions element-wise to the previous states;
    \item the batch of graph invariant values, computed only when required by the agent--environment communication setting; and
    \item the current status of the batch of episodes.
\end{enumerate}
We recognize three possible statuses that an episode may have.
\begin{enumerate}[label=\textbf{(\arabic*)}]
    \item An episode is \emph{in progress} if it is in a state that accepts further actions.
    \item An episode has \emph{terminated} if it has ended due to reaching a terminal state. This status is only possible in RL environments where the tasks are episodic and terminal states exist.
    \item An episode has been \emph{truncated} if it has ended because the required number of steps has been taken. In this case, although the current state is not terminal, no further actions should be performed. This status appears only in RL environments with continuing tasks.
\end{enumerate}
Since all the episodes in any batch are guaranteed to end at the same time, they always have the same status, so we can define the status of a batch of episodes as the status of any of its episodes. This is precisely the status that the \texttt{reset\_batch} and \texttt{step\_batch} methods return. Note that we purposefully distinguish between termination and truncation to keep the API consistent with \texttt{Gymnasium}.

In the context of graph invariant computation, we recognize two types of agent--environment communication settings: sparse and dense. This setting is a configurable parameter of each instance of the \texttt{GraphEnvironment} class and can be reconfigured at any step.
\begin{enumerate}[label=\textbf{(\arabic*)}]
    \item If the sparse setting is selected, then no batch of graph invariant values is computed after each batch of actions, except after the final batch. In that case, the returned values are equal to $f(\varphi(s))$, where $\varphi(s)$ is the batch of underlying graphs corresponding to the final states, and $f$ is a function that accepts a batch of graphs and returns the corresponding graph invariant values.
    
    \item If the dense setting is selected, then a batch of graph invariant values is computed after each batch of actions, and it is equal to $f(\varphi(s))$, where $\varphi(s)$ is the batch of underlying graphs corresponding to the newly obtained states.

    \item Additionally, if the dense setting is selected, then the graph invariant computation may optionally be carried out using a function $\Delta f$, which accepts two batches of graphs of equal size and returns the element-wise differences of the corresponding graph invariant values. In this case, the \texttt{reset\_batch} method computes the graph invariant values using $f$, and afterwards, the \texttt{step\_batch} method incrementally updates them using $\Delta f$.
\end{enumerate}
The idea of optionally using $\Delta f$ in the dense setting is natural, since such a difference function may be more efficient than invoking the original graph invariant function $f$.

Although the methods \texttt{reset\_batch} and \texttt{step\_batch} are not abstract, their behavior is largely determined by the abstract methods \texttt{\_initialize\_batch}, \texttt{\_transition\_batch} and \texttt{state\_batch\_to\_graph\_batch}, all of which must be implemented by any concrete subclass inheriting from \texttt{GraphEnvironment}. The \texttt{reset\_batch} method initializes the starting states by invoking \texttt{\_initialize\_batch} and then computes the graph invariant values if required. Similarly, the \texttt{step\_batch} method performs the state transition through \texttt{\_transition\_batch} and subsequently computes the graph invariant values if required. In both methods, the graph invariant computation involves invoking the abstract method \texttt{state\_batch\_to\_graph\_batch}, which encapsulates $\varphi$ as a pure function. Therefore, concrete subclasses of \texttt{GraphEnvironment} only need to implement the state initialization and state transition logic, without handling the graph invariant computation itself, since this is managed entirely within the \texttt{GraphEnvironment} class. The concrete classes must accept the selected communication setting, the graph invariant function $f$, and optionally the graph invariant difference function $\Delta f$, as constructor arguments, which are then passed to the abstract parent class.

Apart from the three mentioned abstract methods, any concrete class that inherits from \texttt{GraphEnvironment} must also implement the following six abstract properties:
\begin{enumerate}[label=\textbf{(\arabic*)}]
    \item \texttt{state\_length}, which returns the length of the \texttt{NumPy} vectors representing the states;
    \item \texttt{state\_dtype}, which returns the type of the \texttt{NumPy} vectors representing the states;
    \item \texttt{action\_number}, which returns the total number of actions;
    \item \texttt{action\_mask}, which determines what actions are currently available for execution in each of the episodes that are being run in parallel;
    \item \texttt{episode\_length}, which returns the predetermined length of all the episodes that are currently running or are to be run; and
    \item \texttt{is\_continuing}, which determines whether the environment has continuing or episodic RL tasks.
\end{enumerate}
These six properties provide the interface that allows the agent to interact with the environment.

We provide seven classes that inherit from \texttt{GraphEnvironment} and encapsulate nine different RL environments. These environments can be split into three groups:
\begin{enumerate}[label=\textbf{(\arabic*)}]
    \item the linear environments, which are implemented using three separate classes \texttt{LinearBuildEnvironment}, \linebreak \texttt{LinearSetEnvironment} and \texttt{LinearFlipEnvironment};
    \item the global environments, which are implemented using two separate classes \texttt{GlobalSetEnvironment} and \texttt{GlobalFlipEnvironment}; and
    \item the local environments, which are implemented using two separate classes \texttt{LocalSetEnvironment} and \texttt{LocalFlipEnvironment}.
\end{enumerate}
The class naming is largely inspired by the work of Angileri et al.\ \cite{Angileri2025, Angileri2025B}. Moreover, our seven classes are implemented using similar ideas, with the distinction that support is now also provided for directed graphs and graphs with more than two edge colors, and the agent--environment interaction logic is fully encapsulated on the agent side, with the agent treated as a separate entity.

\subsubsection{Linear environments}

The \texttt{LinearBuildEnvironment} class implements the Linear Build environment, which is directly based on Wagner's original approach. This environment models a graph-building game in which the edges (resp.\ arcs) are initially uncolored and are then properly colored one by one, either in row-major or clockwise order. The user can select the graph order $n \ge 2$ and the number of proper edge colors $k \ge 2$, as well as choose whether the graphs should be directed or undirected and whether loops should be allowed. The RL tasks in this environment are episodic, and the episode length equals $\ell$ as given by \eqref{flattened_length_eq}, i.e., the length of either of the two flattened formats with color numbers.

Each state of the Linear Build environment is represented by a binary \texttt{NumPy} vector of length $k \ell$. In this vector, the first $\ell$ bits indicate which of the $\ell$ edges (resp.\ arcs) have been colored with color $1$; the second $\ell$ bits indicate which of the $\ell$ edges (resp.\ arcs) have been colored with color $2$; and so on, up to the $(k - 1)$-th block of $\ell$ bits, where a value of $1$ indicates which of the $\ell$ edges (resp.\ arcs) have been colored with color $k - 1$. The final $\ell$ bits represent a one-hot encoding of the position determining the next edge (resp.\ arc) to be properly colored. In other words, there is either a single value of $1$, whose index determines which edge (resp.\ arc) should be properly colored next, or all values are $0$, indicating a state in which all edges (resp.\ arcs) have been properly colored, i.e., a terminal state. The user can configure whether the edges (resp.\ arcs) should be arranged in row-major or clockwise order. Each action of the Linear Build environment is an integer between $0$ and $k - 1$ that determines which color the next edge (resp.\ arc) should be properly colored with.

The \texttt{LinearSetEnvironment} and \texttt{LinearFlipEnvironment} classes implement the Linear Set and Linear Flip environments, respectively, both of which are inspired by the Linear environment from the framework of Angileri et al. The Linear Set environment functions in exactly the same way as the Linear Build environment, with the difference that the edges (resp.\ arcs) are initially fully colored in some manner, and are then traversed in row-major or clockwise order, and recolored one by one. We believe that this distinction could prove useful while tackling extremal problems where it matters whether the intermediate states correspond to full or partial configurations in some context. The Linear Flip environment is similar to the Linear Set environment, with the difference that the number of proper edge colors is fixed to two, and each action is a binary number that indicates whether the current edge (resp.\ arc) should be flipped or not. More precisely, if the action is $0$, then the color of the current edge (resp.\ arc) should stay the same, and if the action is $1$, then the color should get transformed by the mapping $c \mapsto 1 - c$.

In both the Linear Set and the Linear Flip environment, the user can configure the exact mechanism how the edges (resp.\ arcs) should be initially colored before each of them is recolored in the selected order. This mechanism is encapsulated as a graph generator, i.e., a function that accepts a positive integer and generates a batch of graphs with the corresponding batch size. Every time a batch of episodes is initialized, the configured graph generator is invoked and the generated batch of graphs is used to obtain the initial states. Note that a graph generator may be both deterministic and nondeterministic. The \texttt{environments} package contains four auxiliary functions that help create various graph generators, including the deterministic generator in which all the graphs in the batch are set to the same provided graph.

\begin{example}\label{linear_example}
The following code snippet creates a Linear Build environment that builds $4$-edge-colored looped complete undirected graphs of order three with allowed loops and the edges arranged in clockwise order.
\begin{lstlisting}[language = Python, frame = trBL, escapeinside={(*@}{@*)}, aboveskip=10pt, belowskip=10pt, numbers=left, rulecolor=\color{black}]
def graph_invariant(graph_batch: Graph):
    zero_color_mask = (graph_batch.flattened_row_major_colors == 0).astype(np.float32)
    return np.sum(zero_color_mask, axis=1) ** 2

e1 = LinearBuildEnvironment(
    graph_invariant=graph_invariant,
    graph_order=3,
    flattened_ordering=FlattenedOrdering.CLOCKWISE,
    edge_colors=4,
    allow_loops=True,
)
\end{lstlisting}
Here, the dense communication setting is selected by default and the configured graph invariant is the square of the number of edges colored with color $0$. Assume that our goal is to initialize a batch of four uncolored graphs and then properly color their edges so that the obtained format representations for the adjacency matrix format with color numbers are given by the matrices $\begin{bmatrix} \begin{smallmatrix} 0 & 3 & 1\\ 3 & 0 & 1\\ 1 & 1 & 2 \end{smallmatrix} \end{bmatrix}$, $\begin{bmatrix} \begin{smallmatrix} 0 & 2 & 0\\ 2 & 3 & 2\\ 0 & 2 & 0 \end{smallmatrix} \end{bmatrix}$, $\begin{bmatrix} \begin{smallmatrix} 0 & 1 & 2\\ 1 & 0 & 3\\ 2 & 3 & 0 \end{smallmatrix} \end{bmatrix}$ and $\begin{bmatrix} \begin{smallmatrix} 1 & 3 & 2\\ 3 & 1 & 0\\ 2 & 0 & 1 \end{smallmatrix} \end{bmatrix}$, respectively. To do this, we first initialize the batch of four uncolored graphs as follows.
\begin{lstlisting}[language = Python, frame = trBL, escapeinside={(*@}{@*)}, aboveskip=10pt, belowskip=10pt, numbers=left, rulecolor=\color{black}]
state_batch, graph_invariant_batch, status = e1.reset_batch(4)
\end{lstlisting}
Afterwards, we execute the corresponding batches of actions.
\begin{lstlisting}[language = Python, frame = trBL, escapeinside={(*@}{@*)}, aboveskip=10pt, belowskip=10pt, numbers=left, rulecolor=\color{black}]
state_batch, graph_invariant_batch, status = e1.step_batch(np.array([0, 0, 0, 1], dtype=np.int32))
state_batch, graph_invariant_batch, status = e1.step_batch(np.array([3, 2, 1, 3], dtype=np.int32))
state_batch, graph_invariant_batch, status = e1.step_batch(np.array([0, 3, 0, 1], dtype=np.int32))
state_batch, graph_invariant_batch, status = e1.step_batch(np.array([1, 0, 2, 2], dtype=np.int32))
state_batch, graph_invariant_batch, status = e1.step_batch(np.array([1, 2, 3, 0], dtype=np.int32))
state_batch, graph_invariant_batch, status = e1.step_batch(np.array([2, 0, 0, 1], dtype=np.int32))
\end{lstlisting}
It is not difficult to verify that these actions lead to the desired batch of underlying graphs. The full source for Example \ref{linear_example} can be found in the file \texttt{examples/environment\_examples.py} in \cite{GitHub}. ~\hfill $\Diamond$
\end{example}

\subsubsection{Global environments}

The \texttt{GlobalSetEnvironment} class implements the Global Set environment, which models a graph-building game in which the edges (resp.\ arcs) are initially fully colored in some manner, and then in each step, any edge (resp.\ arc) can be properly recolored with any color. The user can select the graph order $n \ge 2$ and the number of proper edge colors $k \ge 2$, and also choose whether the graphs should be directed or undirected and whether loops should be allowed. Additionally, the user can configure the graph generator that controls how the initial states are obtained every time a batch of episodes is initialized. The RL tasks in this environment are continuing, and the episode length can be selected as a configurable parameter.

Each state of the Global Set environment is represented by a binary \texttt{NumPy} vector of length $(k - 1) \ell$, where $\ell$ is given by \eqref{flattened_length_eq}. In this vector, the first $\ell$ bits indicate which of the edges (resp.\ arcs) are currently colored with color $1$; the second $\ell$ bits indicate which of the edges (resp.\ arcs) are currently colored with color $2$; and so on, up to the last block of $\ell$ bits, where a value of $1$ indicates which of the edges (resp.\ arcs) are currently colored with color $k - 1$. Here, the edges (resp.\ arcs) are assumed to be arranged in row-major or clockwise order, and the user can select which of these two orders should be applied. Each action of the Global Set environment is an integer $a \in \{ 0, 1, 2, \ldots, k \ell - 1 \}$, such that $a \bmod \ell$ signifies the index of the edge (resp.\ arc) that should be properly recolored, while $\lfloor \frac{a}{\ell} \rfloor$ determines which color the chosen edge (resp.\ arc) should be properly recolored with.

We note that our Global Set environment is inspired by the Global and Flip environments from the framework of Angileri et al. The difference is that edges are recolored instead of flipped, providing support for graphs with more than two edge colors. In addition, the \texttt{environments} package contains the \texttt{GlobalFlipEnvironment} class, which implements two variations of the Global Flip environment that directly correspond to the Global and Flip environments from the framework of Angileri et al. These environments are similar to our Global Set environment, with the difference that the number of proper edge colors is fixed to two, and each action indicates whether a selected edge (resp.\ arc) should be flipped or not. The states in the two Global Flip environments are represented in the same way as in the Global Set environment.

\pagebreak
The action spaces of the two Global Flip environments are not the same. The user can select one of the two environment variations by configuring the boolean \texttt{flip\_only} parameter. If \texttt{flip\_only} is set to \texttt{False}, then each action is an integer $a \in \{ 0, 1, 2, \ldots, 2\ell - 1\}$, such that $a \bmod \ell$ signifies the index of the selected edge (resp.\ arc), while $\lfloor \frac{a}{\ell} \rfloor$ is a binary number that indicates whether the selected edge (resp.\ arc) should be flipped. In this case, the selected edge (resp.\ arc) does not necessarily have to be flipped. On the other hand, if the \texttt{flip\_only} parameter is set to \texttt{True}, then each action is an integer from $\{ 0, 1, 2, \ldots, \ell - 1 \}$ that signifies the index of the edge (resp.\ arc) to be flipped. Here, one edge (resp.\ arc) is selected in each step and it must be flipped. Therefore, the variation of our Global Flip environment where \texttt{flip\_only} is \texttt{False} corresponds to the Global environment from the framework of Angileri et al., while the variation where \texttt{flip\_only} is \texttt{True} corresponds to the Flip environment from the same previous framework.

\begin{example}\label{global_example}
Assume that our goal is to create a Global Flip environment that builds $2$-edge-colored looped complete undirected graphs of order five and without allowed loops. In addition, we want to arrange the edges in row-major order and to enforce edge flipping in each step. Such an environment can be instantiated using the following code snippet.
\begin{lstlisting}[language = Python, frame = trBL, escapeinside={(*@}{@*)}, aboveskip=10pt, belowskip=10pt, numbers=left, rulecolor=\color{black}]
def graph_invariant(graph_batch: Graph):
    degrees = np.sum(graph_batch.adjacency_matrix_colors, axis=2)
    return np.sum(degrees**2, axis=1).astype(np.float32)

e2 = GlobalFlipEnvironment(
    graph_invariant=graph_invariant,
    graph_order=5,
    episode_length=4,
    flip_only=True,
    flattened_ordering=FlattenedOrdering.ROW_MAJOR,
    initial_graph_generator=create_fixed_graph_generator(
        fixed_graph=MonochromaticGraph(
            graph_formats={GraphFormat.FLATTENED_ROW_MAJOR_COLORS},
            graph_order=5,
            selected_color=1,
        ),
        graph_format=GraphFormat.FLATTENED_ROW_MAJOR_COLORS,
    ),
    sparse_setting=True,
)
\end{lstlisting}
Here, we have selected the sparse communication setting and configured the graph invariant to be the sum of squares of all vertex degrees. We have also configured the episode length to four and ensured that the all edges in each initial graph are colored with color $1$. The latter was achieved by using the auxiliary \texttt{create\_fixed\_graph\_generator} function together with the \texttt{MonochromaticGraph} class, which instantiates graphs where all the edges (resp.\ arcs) are colored with the same color. We can then initialize two parallel episodes as follows.
\begin{lstlisting}[language = Python, frame = trBL, escapeinside={(*@}{@*)}, aboveskip=10pt, belowskip=10pt, numbers=left, rulecolor=\color{black}]
state_batch, graph_invariant_batch, status = e2.reset_batch(2)
\end{lstlisting}
Now, suppose we execute the following batches of actions.
\begin{lstlisting}[language = Python, frame = trBL, escapeinside={(*@}{@*)}, aboveskip=10pt, belowskip=10pt, numbers=left, rulecolor=\color{black}]
state_batch, graph_invariant_batch, status = e2.step_batch(np.array([0, 2], dtype=np.int32))
state_batch, graph_invariant_batch, status = e2.step_batch(np.array([1, 7], dtype=np.int32))
state_batch, graph_invariant_batch, status = e2.step_batch(np.array([5, 1], dtype=np.int32))
state_batch, graph_invariant_batch, status = e2.step_batch(np.array([9, 7], dtype=np.int32))
\end{lstlisting}
Executing the above code snippet truncates the episodes and leads to the underlying graphs whose format representations for the adjacency matrix format with color numbers are given by $\begin{bmatrix}\begin{smallmatrix} 0 & 0 & 0 & 1 & 1\\ 0 & 0 & 1 & 0 & 1\\ 0 & 1 & 0 & 1 & 1\\ 1 & 0 & 1 & 0 & 0\\ 1 & 1 & 1 & 0 & 0 \end{smallmatrix}\end{bmatrix}$ and $\begin{bmatrix}\begin{smallmatrix} 0 & 1 & 0 & 0 & 1\\ 1 & 0 & 1 & 1 & 1\\ 0 & 1 & 0 & 1 & 1\\ 0 & 1 & 1 & 0 & 1\\ 1 & 1 & 1 & 1 & 0 \end{smallmatrix}\end{bmatrix}$, respectively. The full source code for Example \ref{global_example} is available in the file \texttt{examples/environment\_examples.py} in \cite{GitHub}. ~\hfill $\Diamond$
\end{example}

\subsubsection{Local environments}

The \texttt{LocalSetEnvironment} and \texttt{LocalFlipEnvironment} classes implement the Local Set environment and two variations of the Local Flip environment, respectively, all of which are inspired by the Local environment from the framework of Angileri et al. The Local Set environment models a graph-building game in which the edges (resp.\ arcs) are initially fully colored in some manner, and the agent moves from one vertex to another according to a chosen strategy, thereby traversing the existing edges (resp.\ arcs) and properly recoloring them. More precisely, in each step, the agent is located at a vertex and must select an edge incident to this vertex or an arc starting at this vertex, then traverse it and move to the other endpoint of the traversed edge (resp.\ arc). While traversing an edge (resp.\ arc), the agent also properly recolors it with a selected color.

As in the previous environments, the user can select the graph order $n \ge 2$ and the number of proper edge colors $k \ge 2$, as well as choose whether the graphs should be directed or undirected and whether loops should be allowed. Additionally, the user can configure the graph generator that controls how the initial states are obtained when a batch of episodes is initialized. The user can also select the vertex at which the agent should start the recoloring procedure. The RL tasks in the Local Set environment are continuing, with the episode length being a configurable parameter.

Each state of the Local Set environment is represented by a binary \texttt{NumPy} vector of length $(k - 1) \ell + n$, where $\ell$ is given by \eqref{flattened_length_eq}. In this vector, the first $(k - 1) \ell$ bits have the same meaning as in the global and linear environments. Once again, the edges (resp.\ arcs) are assumed to be arranged in row-major or clockwise order, and the user can select which of these two orders should be applied. The final $n$ bits of the state vector represent a one-hot encoding of the position determining the vertex where the agent is currently located. In other words, there is a single value of $1$ whose index determines the vertex where the agent is located. Each action of the Local Set environment is an integer $a \in \{ 0, 1, 2, \ldots, k n - 1 \}$, such that $a \bmod n$ signifies the vertex that the agent should move to from the current vertex, while $\lfloor \frac{a}{n} \rfloor$ determines which color the traversed edge (resp.\ arc) should be properly recolored with. Note that, unlike in the previous environments where any action is available for execution in any non-terminal state, it is prohibited to move from a vertex to itself if loops are not allowed.

The two variations of the Local Flip environment function in exactly the same way as the Local Set environment, with the difference that the number of proper edge colors is fixed to two, and each action indicates whether a traversed edge (resp.\ arc) should be flipped or not. The states in the two Local Flip environments are represented in the same way as in the Local Set environment. The action spaces of the two Local Flip environments are not the same, and the user can select one of the two environment variations by configuring the boolean \texttt{flip\_only} parameter, similarly to the Global Flip environments. If \texttt{flip\_only} is set to \texttt{False}, then each action is an integer $a \in \{ 0, 1, 2, \ldots, 2n - 1 \}$, such that $a \bmod n$ signifies the vertex that the agent should move to from the current vertex, while $\lfloor \frac{a}{n} \rfloor$ is a binary number that indicates whether the traversed edge (resp.\ arc) should be flipped. On the other hand, if \texttt{flip\_only} is set to \texttt{True}, then each action is an integer from $\{ 0, 1, 2, \ldots, n - 1 \}$ that signifies the vertex that the agent should move to, with the traversed edge (resp.\ arc) being necessarily flipped.

\begin{example}\label{local_example}
The following code snippet creates a Local Set environment that builds $3$-edge-colored looped complete directed graphs of order four and without allowed loops, such that the agent starts the recoloring procedure at vertex $0$.
\begin{lstlisting}[language = Python, frame = trBL, escapeinside={(*@}{@*)}, aboveskip=10pt, belowskip=10pt, numbers=left, rulecolor=\color{black}]
def graph_invariant(graph_batch: Graph):
    adj_1 = graph_batch.adjacency_matrix_binary[:, -2, :, :]
    trace_sum_1 = np.trace(adj_1 @ adj_1 @ adj_1, axis1=1, axis2=2)

    adj_2 = graph_batch.adjacency_matrix_binary[:, -1, :, :]
    trace_sum_2 = np.trace(adj_2 @ adj_2 @ adj_2, axis1=1, axis2=2)

    return (trace_sum_1 + trace_sum_2).astype(np.float32) / 3.0

e3 = LocalSetEnvironment(
    graph_invariant=graph_invariant,
    graph_order=4,
    episode_length=6,
    flattened_ordering=FlattenedOrdering.ROW_MAJOR,
    edge_colors=3,
    is_directed=True,
    starting_vertex=0,
)
\end{lstlisting}
\pagebreak
We have also selected the dense communication setting by default and configured the graph invariant to be the combined number of $1$- and $2$-monochromatic directed cycles of length three. In addition, the arcs have been arranged in row-major order and the episode length has been configured to six. We can now initialize a single episode as follows.
\begin{lstlisting}[language = Python, frame = trBL, escapeinside={(*@}{@*)}, aboveskip=10pt, belowskip=10pt, numbers=left, rulecolor=\color{black}]
state_batch, graph_invariant_batch, status = e3.reset_batch(1)
\end{lstlisting}
Assume that our goal is to traverse the directed walk $(0, 2, 3, 0, 1, 3, 0)$, so that all the traversed arcs are colored with color $1$, apart from the last one, which is colored with color $2$. This can be achieved through the following code snippet.
\begin{lstlisting}[language = Python, frame = trBL, escapeinside={(*@}{@*)}, aboveskip=10pt, belowskip=10pt, numbers=left, rulecolor=\color{black}]
state_batch, graph_invariant_batch, status = e3.step_batch(np.array([6], dtype=np.int32))
state_batch, graph_invariant_batch, status = e3.step_batch(np.array([7], dtype=np.int32))
state_batch, graph_invariant_batch, status = e3.step_batch(np.array([4], dtype=np.int32))
state_batch, graph_invariant_batch, status = e3.step_batch(np.array([5], dtype=np.int32))
state_batch, graph_invariant_batch, status = e3.step_batch(np.array([7], dtype=np.int32))
state_batch, graph_invariant_batch, status = e3.step_batch(np.array([8], dtype=np.int32))
\end{lstlisting}
As it turns out, the combined number of $1$- and $2$-monochromatic directed cycles of length three starts at $0$, then becomes $1$ in the third step, then reaches $2$ in the fifth step, and finally returns to $0$ in the final step. This can easily be verified by executing the full source code for Example \ref{local_example}, which can be found in the file \texttt{examples/environment\_examples.py} in \cite{GitHub}. ~\hfill $\Diamond$
\end{example}

\subsection{Agents and RL methods}

The \texttt{agents} package contains the abstract class \texttt{GraphAgent}, which encapsulates RL agents in extremal graph theory applications, together with three concrete classes that inherit from this class. The main methods of the \texttt{GraphAgent} class are \texttt{reset}, which initializes (or reinitializes) the agent and prepares it to start the learning process, and \texttt{step}, which performs a single iteration of the learning process. Both of these methods are abstract and need to be implemented by any concrete subclass of \texttt{GraphAgent}. The method names follow the same naming pattern as the primary methods of \texttt{GraphEnvironment}. In addition, any concrete class that inherits from \texttt{GraphAgent} must also implement the following three abstract properties:
\begin{enumerate}[label=\textbf{(\arabic*)}]
    \item \texttt{step\_count}, which returns the number of executed learning iterations;
    \item \texttt{best\_score}, which returns the best value of the target graph invariant achieved so far; and
    \item \texttt{best\_graph}, which returns a graph attaining the best achieved graph invariant value.
\end{enumerate}

We provide three different RL methods: the Deep Cross-Entropy, the REINFORCE, and the PPO method, implemented using \texttt{PyTorch} in the \texttt{DeepCrossEntropyAgent}, \texttt{ReinforceAgent}, and \texttt{PPOAgent} classes, respectively. All of these classes inherit from \texttt{GraphAgent} and implement an RL agent that interacts with a configurable instance of \texttt{GraphEnvironment} by playing the graph-building game induced by the environment, thereby generating graphs.

The \texttt{DeepCrossEntropyAgent} class implements an RL agent using the Deep Cross-Entropy method. In each iteration of the learning process, the agent plays a predetermined number of parallel graph-building games and executes batches of actions according to a strategy modeled by a policy network. The sparse communication setting is enforced, and the graph invariant value is computed for the underlying graph of the final state of each episode. Afterwards, a certain number of episodes with the highest graph invariant values are used to train the policy network with cross-entropy loss, while another subset of top-performing episodes is carried over to the next generation. This completes one iteration of the learning process. The user can select the number of parallel episodes to be run in each generation, the number of top-performing episodes used for training the policy network, and the number of top-performing episodes carried over to the next generation. In addition, it is possible to configure the policy network itself, as well as the optimizer responsible for updating its parameters.

The \texttt{ReinforceAgent} class encapsulates an RL agent that uses the REINFORCE method. In each iteration of the learning process, the agent plays a configured number of parallel graph-building games and selects actions according to a strategy modeled by a policy network, as in the \texttt{DeepCrossEntropyAgent} class. Unlike the \texttt{DeepCrossEntropyAgent} class, the \texttt{ReinforceAgent} class enforces the dense communication setting and computes both the final graph invariant values and the discounted returns at each step for all episodes run in parallel. While computing the discounted returns, the reward is naturally defined as the increase between consecutive graph invariant values. Afterwards, a specified number of episodes with the highest graph invariant values are used to train the policy network according to the REINFORCE algorithm, which completes one iteration of the learning process. As in the \texttt{DeepCrossEntropyAgent} class, the user can select the number of parallel episodes to be run in each generation and the number of top-performing episodes used for training the policy network. Additionally, it is possible to configure the policy network, the optimizer responsible for training it, the discount factor used when computing the discounted returns, and whether a baseline should be applied in the training process to reduce variance.

The \texttt{PPOAgent} class implements an RL agent using the PPO method. Similar to the \texttt{ReinforceAgent} class, the agent generates a configured number of graphs per learning iteration and computes the corresponding graph invariant values and discounted returns, with the reward defined as the increase between consecutive graph invariant values. Unlike the \texttt{ReinforceAgent} class, two neural networks are employed: the policy network, which models the strategy used to play the graph-building game, and the value network, which estimates the desirability of a given state. A specified number of episodes with the highest graph invariant values are used to train both networks according to the PPO algorithm, which completes one iteration of the learning process. Most of the parameters configurable in the \texttt{ReinforceAgent} class are also available in the \texttt{PPOAgent} class, together with some additional parameters specific to the PPO algorithm. For instance, the user can choose the number of epochs executed in each learning iteration, the clamping coefficient used when computing the policy loss in each epoch, as well as the coefficient that scales the value loss when computing the total loss. For more details, the reader can refer to the framework documentation \cite{Documentation}.

For the sake of better exploration, all three concrete classes inheriting from \texttt{GraphAgent} support the execution of random actions. More precisely, at each step there is a given probability of the action issued by the policy being ignored and a randomly chosen action being performed instead. In this case, the computation of the random action probability is governed by a configurable mechanism encapsulated in the abstract class \texttt{RandomActionMechanism}. We offer three concrete classes that inherit from this class and exhibit different behaviors for controlling the random action probability. By default, the random action probability is set to zero, i.e., no random actions are executed.

\begin{figure}[H]
\centering
\subcaptionbox{The best score versus step count plot.\label{plot_1}}[0.99\textwidth]
{
    \centering
    \includegraphics[width=0.99\textwidth]{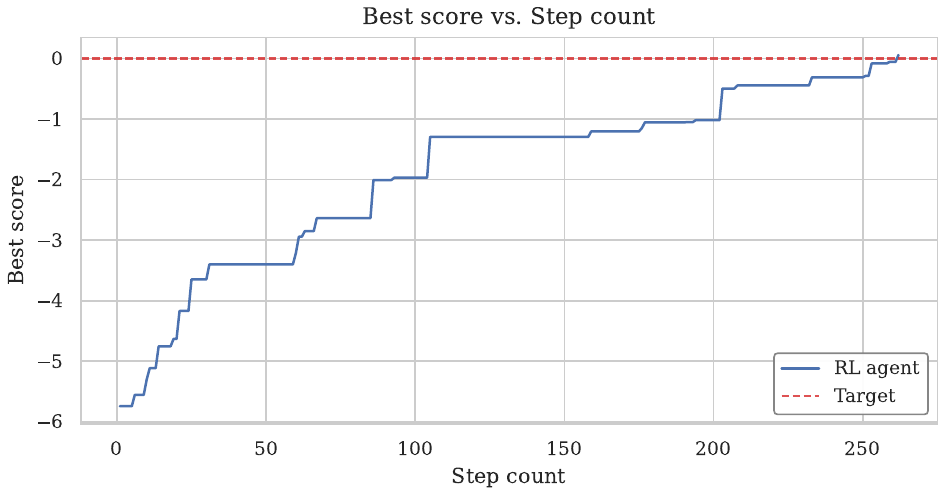}
}
\\
\subcaptionbox{A counterexample of order $16$ to Conjecture~\ref{agent_conj}.\label{counter_1b}}[0.44\textwidth]
{
    \centering
    \includegraphics[width=0.39\textwidth]{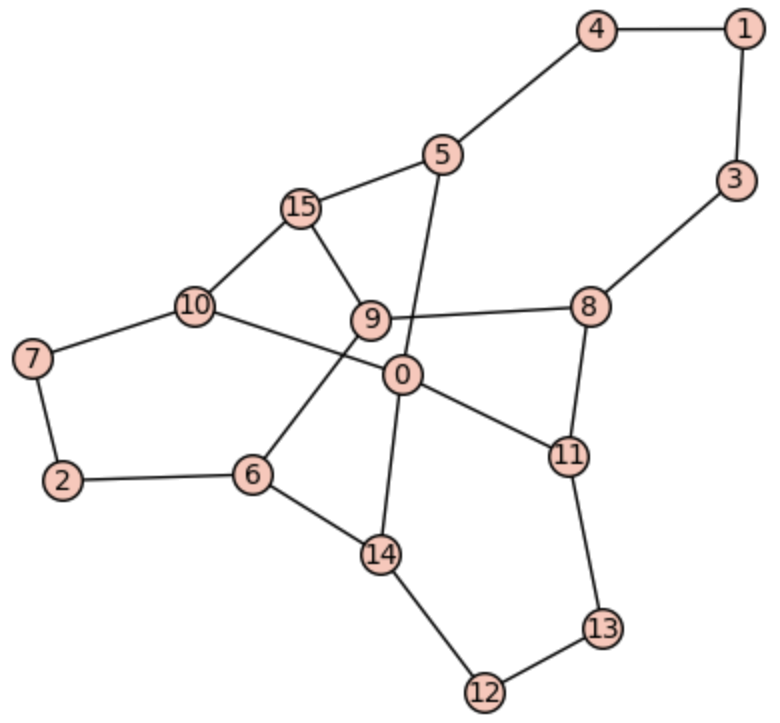}
}
\qquad
\subcaptionbox{Another counterexample of order $16$ to Conjecture~\ref{agent_conj}.\label{counter_1c}}[0.50\textwidth]
{
    \centering
    \includegraphics[width=0.50\textwidth]{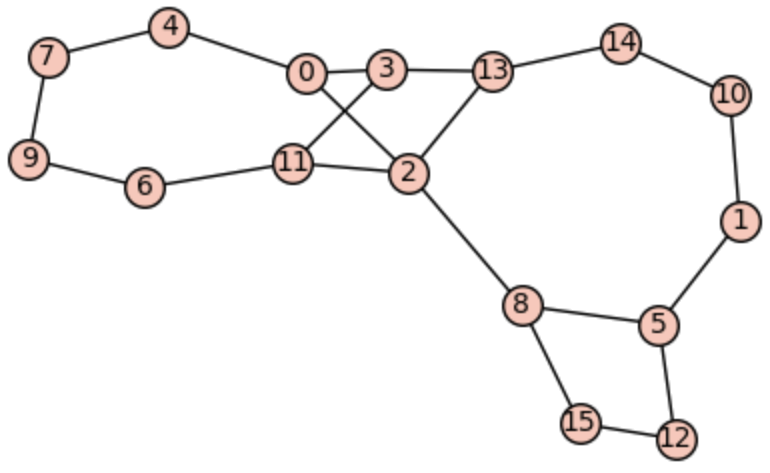}
}
\caption{The best score versus step count plot for the Deep Cross-Entropy agent based disproof of Conjecture~\ref{agent_conj}, and two counterexamples of order $16$.}
\end{figure}

\begin{example}\label{agent_example}
As already mentioned, Ghebleh et al.\ \cite{GheYaKaSte2024} used their reimplementation of Wagner's approach to disprove many of the upper bounds on the Laplacian spectral radius previously conjectured in \cite{BraHaSte2006}. Among the refuted upper bounds, one was the following.

\begin{conjecture}[\hspace{1sp}{\cite[Upper bound 3]{GheYaKaSte2024}}]\label{agent_conj}
    For any nontrivial connected simple graph $G$, we have
    \[
        \mu(G) \le \max_{v \in V(G)} \left( \frac{m(v)^2}{d(v)} + m(v) \right),
    \]
    where $\mu(G)$ denotes the Laplacian spectral radius of graph $G$, $d(v)$ is the degree of a vertex $v$ in $G$, and $m(v)$ is the average degree of all the neighbors of a vertex $v$ in $G$.
\end{conjecture}

We now demonstrate how the RLGT framework can be applied to concisely disprove Conjecture~\ref{agent_conj}. To begin, we need to implement the graph invariant function, which in our situation is
\[
    G \mapsto \mu(G) - \max_{v \in V(G)} \left( \frac{m(v)^2}{d(v)} + m(v) \right) .
\]
Although there are more efficient or faster ways to do this, the following code snippet is sufficient for our example.

\begin{lstlisting}[language = Python, frame = trBL, escapeinside={(*@}{@*)}, aboveskip=10pt, belowskip=10pt, numbers=left, rulecolor=\color{black}]
def graph_invariant(graph_batch: Graph) -> np.ndarray:
    adjacency_matrix_batch = graph_batch.adjacency_matrix_colors.astype(np.float64)

    d_batch = adjacency_matrix_batch.sum(axis=2)
    d_batch_fixed = np.maximum(d_batch, 1)
    m_batch = adjacency_matrix_batch @ d_batch[..., None]
    m_batch = m_batch[..., 0] / d_batch_fixed
    m_batch_fixed = np.maximum(m_batch, 1)

    laplacian_matrix_batch = -adjacency_matrix_batch
    index_range = np.arange(adjacency_matrix_batch.shape[1])
    laplacian_matrix_batch[:, index_range, index_range] += d_batch
    spectrum_batch = np.linalg.eigvalsh(laplacian_matrix_batch)
    mu_batch = spectrum_batch[:, -1]

    right_hand_side_batch = np.max(m_batch_fixed**2 / d_batch_fixed + m_batch_fixed, axis=1)
    result = mu_batch - right_hand_side_batch

    temp = graph_batch.adjacency_matrix_colors.astype(bool) | np.eye(
        graph_batch.graph_order, dtype=bool
    )
    power = 1
    while power < graph_batch.graph_order - 1:
        temp = (temp @ temp).astype(bool)
        power *= 2

    result[~np.all(temp[:, 0, :], axis=1)] = -10.0

    return result.astype(np.float32)
\end{lstlisting}

Note that the function above assigns a graph invariant value of $-10$ to any disconnected graph, since the conjecture applies only to connected graphs. With the graph invariant function at our disposal, the entire conjecture disproval can be carried out through the following short code snippet.

\begin{lstlisting}[language = Python, frame = trBL, escapeinside={(*@}{@*)}, aboveskip=10pt, belowskip=10pt, numbers=left, rulecolor=\color{black}]
def a1_example(graph_order: int):
    policy_network = nn.Sequential(
        nn.Linear(graph_order * (graph_order - 1), 72),
        nn.ReLU(),
        nn.Dropout(0.2),
        nn.Linear(72, 12),
        nn.ReLU(),
        nn.Dropout(0.2),
        nn.Linear(12, 2),
    )

    agent = DeepCrossEntropyAgent(
        environment=LinearBuildEnvironment(
            graph_invariant=graph_invariant,
            graph_order=graph_order,
        ),
        policy_network=policy_network,
        optimizer=optim.Adam(policy_network.parameters(), lr=0.003),
    )

    print("Deep Cross-Entropy agent + Linear Build environment")
    print("Starting...")
    agent.reset()

    while True:
        agent.step()
        print(f"Learning iterations: {agent.step_count}. Best score: {agent.best_score:.3f}.")

        if agent.best_score > 0.0001:
            print("Success! The following graph is a solution:")
            print(agent.best_graph.adjacency_matrix_colors)

            break

        if agent.step_count >= 1000:
            print("Restarting...")
            agent.reset()


if __name__ == "__main__":
    a1_example(graph_order=16)
\end{lstlisting}
In the above code, the Deep Cross-Entropy agent is used in conjunction with the Linear Build environment to efficiently find a counterexample of order $16$ to Conjecture \ref{agent_conj}. All default arguments are used, and the policy network architecture is the same as in \cite{GheYaKaSte2024}. The corresponding best score versus step count plot appears in Figure \ref{plot_1}, while two obtained counterexample graphs of order $16$ are shown in Figures \ref{counter_1b} and \ref{counter_1c}. The full source code for Example \ref{agent_example} is available in the file \texttt{examples/agent\_examples.py} in \cite{GitHub}. In this file, there are two more agent--environment combinations capable of refuting Conjecture \ref{agent_conj}: the REINFORCE agent together with the Global Flip environment with enforced edge flipping, and the PPO agent together with the Local Set environment. ~\hfill $\Diamond$
\end{example}

\section{Applications}\label{sc_applications}

In this section, we present three applications of the RLGT framework to concrete extremal graph theory problems, illustrating the framework's efficiency and expressive power. The first application extends Example \ref{agent_example} to all the conjectures disproved in \cite{GheYaKaSte2024}, while the other two address unrelated problems. The full source code and execution results for these applications are available in the \texttt{applications} folder in \cite{GitHub}.

\subsection{Laplacian spectral radius}\label{appl_1}

The \emph{Laplacian spectral radius} of a graph $G$, denoted by $\mu(G)$, is the largest eigenvalue of the Laplacian matrix of $G$. While reimplementing Wagner's approach, Ghebleh et al.\ \cite{GheYaKaSte2024} investigated 68 upper bounds on the Laplacian spectral radius previously conjectured in \cite{BraHaSte2006}, all of which have the form
\[
    \mu(G) \le \max_{v \in V(G)} h(d(v), m(v)) \qquad \mbox{or} \qquad \mu(G) \le \max_{uv \in E(G)} h(d(u), m(u), d(v), m(v)) ,
\]
where $d(v)$ denotes the degree of a vertex $v$ in the graph $G$, $m(v)$ denotes the average degree of all the neighbors of a vertex $v$ in $G$, and $h$ is some configurable real function. Note that in the latter case, $h$ must be symmetric with respect to $u$ and $v$. Using their newly implemented RL framework, Ghebleh et al.\ successfully disproved 25 of these conjectured inequalities, while five more were refuted via exhaustive search without RL.

Conjecture \ref{agent_conj}, which we disproved in Example \ref{agent_example}, is actually Upper bound 3 from \cite[Appendix~B]{GheYaKaSte2024}. We now demonstrate the efficiency of the RLGT framework by replicating the RL-based disproofs from \cite{GheYaKaSte2024}, i.e., by obtaining new counterexamples to all the inequalities previously refuted via RL. This can be achieved by reworking the graph invariant function from Example \ref{agent_example} as follows.
\begin{lstlisting}[language = Python, frame = trBL, escapeinside={(*@}{@*)}, aboveskip=10pt, belowskip=10pt, numbers=left, rulecolor=\color{black}]
def compute_graph_invariant(graph_batch: Graph, expression_index: int) -> np.ndarray:
    adjacency_matrix_batch = graph_batch.adjacency_matrix_colors.astype(np.float64)

    d_batch = adjacency_matrix_batch.sum(axis=2)
    d_batch_fixed = np.maximum(d_batch, 1)
    m_batch = adjacency_matrix_batch @ d_batch[..., None]
    m_batch = m_batch[..., 0] / d_batch_fixed
    m_batch_fixed = np.maximum(m_batch, 1)

    laplacian_matrix_batch = -adjacency_matrix_batch
    index_range = np.arange(adjacency_matrix_batch.shape[1])
    laplacian_matrix_batch[:, index_range, index_range] += d_batch
    spectrum_batch = np.linalg.eigvalsh(laplacian_matrix_batch)
    mu_batch = spectrum_batch[:, -1]

    if expression_index <= 32:
        right_hand_side_batch = np.max(
            LAPLACIAN_EXPRESSIONS[expression_index](d_batch_fixed, m_batch_fixed), axis=1
        )
    else:
        b, u, v = np.nonzero(np.triu(graph_batch.adjacency_matrix_colors, k=1))

        du = d_batch_fixed[b, u]
        mu = m_batch_fixed[b, u]
        dv = d_batch_fixed[b, v]
        mv = m_batch_fixed[b, v]

        all_right_hand_sides = LAPLACIAN_EXPRESSIONS[expression_index](du, mu, dv, mv)
        np.nan_to_num(all_right_hand_sides, nan=-1000.0, copy=False)
        right_hand_side_batch = np.full(graph_batch.batch_size, -np.inf)
        np.maximum.at(right_hand_side_batch, b, all_right_hand_sides)

    result = mu_batch - right_hand_side_batch

    temp = graph_batch.adjacency_matrix_colors.astype(bool) | np.eye(
        graph_batch.graph_order, dtype=bool
    )
    power = 1
    while power < graph_batch.graph_order - 1:
        temp = (temp @ temp).astype(bool)
        power *= 2

    result[~np.all(temp[:, 0, :], axis=1)] = -10.0

    return result.astype(np.float32)
\end{lstlisting}

Here, \texttt{LAPLACIAN\_EXPRESSIONS} is a global dictionary containing all the right-hand side functions $h$. By configuring the agent and the environment in largely the same manner, it is indeed possible to replicate the results from \cite{GheYaKaSte2024}. The full source code of this script is provided in the \texttt{auto\_laplacian\_solver.py} file, while the obtained counterexample graphs are given in the \texttt{auto\_laplacian\_solutions.txt} file in the bitmask format, with each line corresponding to a separate graph. The \texttt{auto\_laplacian\_checker.py} \texttt{SageMath} \cite{SageMath} script can now conveniently be used to verify the validity of these counterexamples and enumerate all the conjectured inequalities that have been disproved. For more details, the reader can refer to \cite{GitHub}.

\subsection{Graph energy and matching number}\label{appl_2}

Before we proceed, we need some additional definitions and notation. The \emph{energy} of a simple graph $G$, denoted by $\mathcal{E}(G)$, is the sum of absolute values of all the eigenvalues of $A(G)$, as introduced by Gutman \cite{Gutman1978} in 1978. Also, the \emph{matching number} of a simple graph $G$, denoted by $\nu(G)$, is the size of a maximum matching in $G$. Finally, let $\Delta(G)$ denote the maximum vertex degree in a graph $G$. In a recent paper, Akbari, Alazemi and Anđelić investigated the relationship between the graph energy and matching number, proving the following theorem.

\begin{theorem}[\hspace{1sp}{\cite[Theorem~18]{AkAlaAn2021}}]\label{aaa_th}
For any connected graph $G$ with $\Delta(G) \ge 6$, we have $\mathcal{E}(G) \le 2 \nu(G) \sqrt{\Delta(G)}$.
\end{theorem}

\noindent
It is natural to ask whether the conditions from Theorem \ref{aaa_th} can be relaxed, leading to the following conjecture.

\begin{conjecture}[\hspace{1sp}{\cite[Conjecture~23]{AkAlaAn2021}}]\label{aaa_conj}
For any connected graph $G \not\cong C_3, C_5, C_7$ with $\Delta(G) \in \{ 2, 3, 4, 5 \}$, we have $\mathcal{E}(G) \le 2 \nu(G) \sqrt{\Delta(G)}$.
\end{conjecture}

Using Wagner's original approach, Conjecture \ref{aaa_conj} was recently refuted \cite{SteDaSte2021} through an infinite number of counterexamples. The structural patterns of these counterexample graphs were uncovered via RL, allowing the manual construction of two infinite families of counterexamples. We now demonstrate how Conjecture \ref{aaa_conj} can be easily refuted by applying the RLGT framework together with \texttt{SageMath}, which is useful for computing the matching numbers. The graph invariant function $G \mapsto \mathcal{E}(G) - 2 \nu(G) \sqrt{\Delta(G)}$ can be implemented using the \texttt{SageMath} features as follows.

\begin{lstlisting}[language = Python, frame = trBL, escapeinside={(*@}{@*)}, aboveskip=10pt, belowskip=10pt, numbers=left, rulecolor=\color{black}]
def graph_invariant(graph_batch) -> np.ndarray:
    scores = np.empty(graph_batch.batch_size, dtype=np.float32)

    for index in range(graph_batch.batch_size):
        g = Graph(matrix(graph_batch.adjacency_matrix_colors[index]))
        if not g.is_connected():
            scores[index] = -2000.0
            continue

        delta = max(g.degree())
        if delta > 5:
            scores[index] = -2000.0
            continue

        nu = len(g.matching())
        eigenvalues = g.adjacency_matrix().eigenvalues()
        energy = sum(abs(eigenvalue) for eigenvalue in eigenvalues)

        scores[index] = energy - 2 * nu * sqrt(delta)

    return scores
\end{lstlisting}

\begin{figure}[t]
\centering
\subcaptionbox{The best score versus step count plot.\label{plot_2}}[0.99\textwidth]
{
    \centering
    \includegraphics[width=0.99\textwidth]{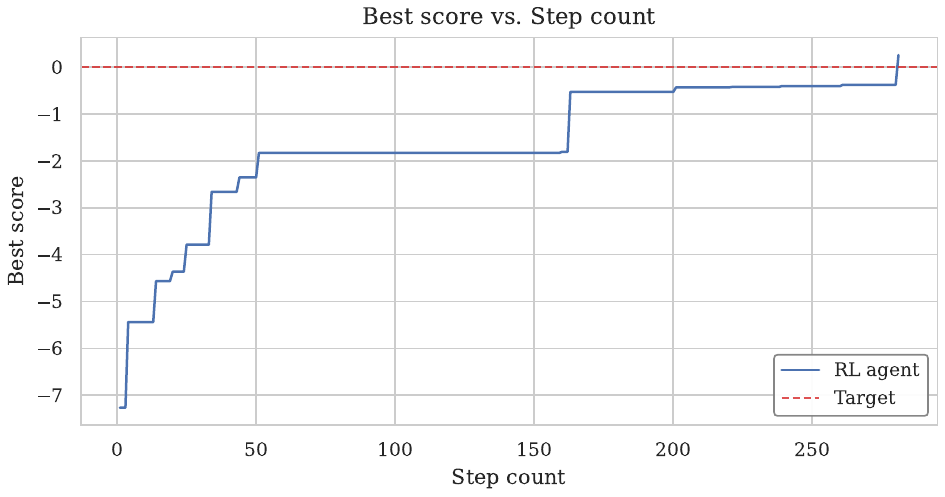}
}
\\
\subcaptionbox{A counterexample of order $14$ to Conjecture~\ref{aaa_conj}.\label{counter_2b}}[0.46\textwidth]
{
    \centering
    \includegraphics[width=0.46\textwidth]{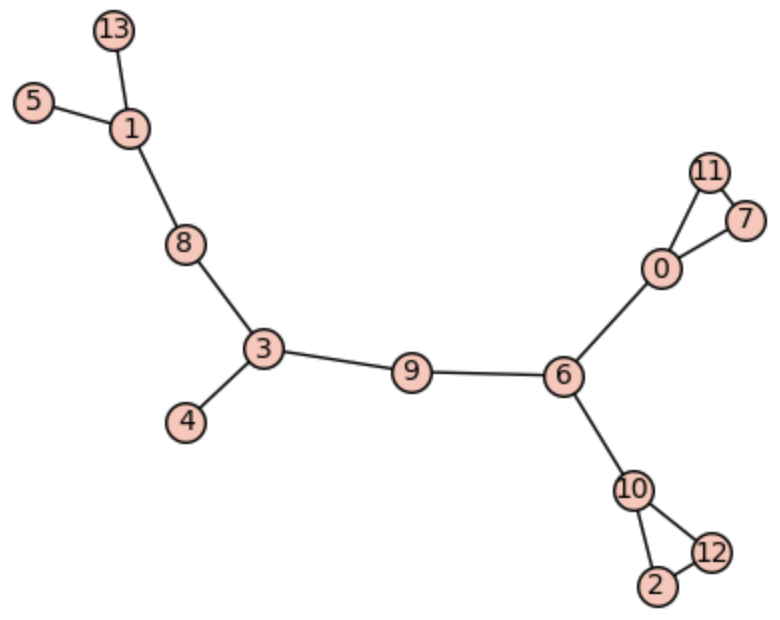}
}
\qquad
\subcaptionbox{Another counterexample of order $14$ to Conjecture~\ref{aaa_conj}.\label{counter_2c}}[0.48\textwidth]
{
    \centering
    \includegraphics[width=0.46\textwidth]{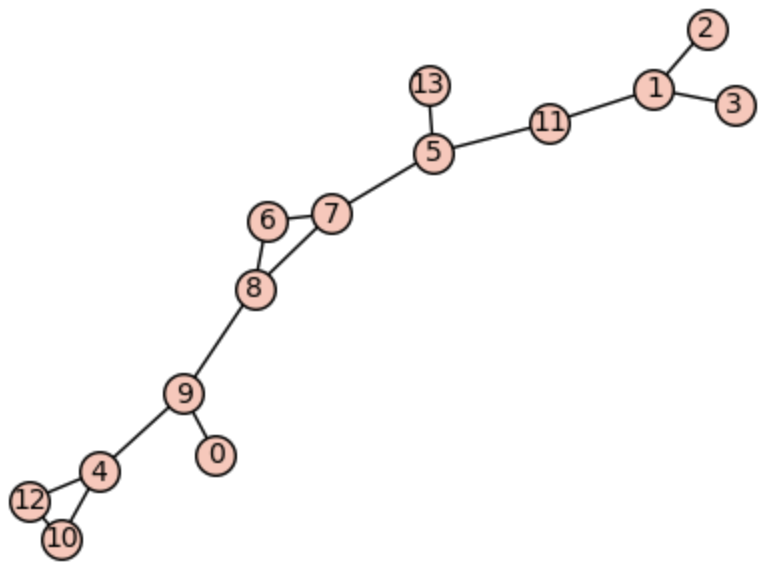}
}
\caption{The best score versus step count plot for the disproof of Conjecture~\ref{aaa_conj}, and two counterexamples of order $14$.}
\end{figure}

The above function assigns a graph invariant value of $-2000$ to any disconnected graph or graph whose maximum vertex degree is at least six. Using this function together with the Deep Cross-Entropy agent in conjunction with the Linear Build environment, we can easily disprove Conjecture \ref{aaa_conj} by finding counterexamples of order $14$. The corresponding best score versus step count plot appears in Figure \ref{plot_2}, while two obtained counterexample graphs of order $14$ are shown in Figures \ref{counter_2b} and \ref{counter_2c}. Observe that these two counterexamples resemble the graphs from one of the two infinite families of counterexamples constructed in \cite{SteDaSte2021}, highlighting the positive outcome of the agent learning process.

The full source code for the conjecture solving \texttt{SageMath} script is provided in the \texttt{wine\_glasses\_solver.py} file, while four obtained counterexamples are given in the \texttt{wine\_glasses\_solutions.txt} file in the same format as in Subsection \ref{appl_1}. The validity of the obtained counterexamples can easily be verified by running the \texttt{wine\_glasses\_checker.py} \texttt{SageMath} script.

\subsection{Mostar index}

We end the section with one non-application, i.e., an unsuccessful attempt to refute a conjecture. The \emph{Mostar index} of a connected simple graph $G$, denoted by $\mathrm{Mo}(G)$, is defined as
\[
    \mathrm{Mo}(G) \coloneqq \sum_{uv \in E(G)} \left| n_G(u, v) - n_G(v, u) \right|,
\]
where $n_G(u, v)$ is the number of vertices in $G$ closer to $u$ than to $v$ and $n_G(v, u)$ is defined analogously, as recently introduced by Došlić et al.\ \cite{DoMaSkrTiZu2018} and independently discovered by Sharafdini and Réti \cite{ShaRe2020}. Also, let $G_1 \lor G_2$ denote the join of two graphs $G_1$ and $G_2$, i.e., the graph that arises by taking two disjoint copies of $G_1$ and $G_2$ and adding all the edges with one endpoint in $G_1$ and the other in $G_2$. Despite being natural, the following conjecture on the extremality of the Mostar index still seems to be open.

\begin{conjecture}[\hspace{1sp}{\cite{AliDo2021, DoMaSkrTiZu2018}}]\label{mostar_conj}
    For any $n \ge 3$, the graph $K_{\lfloor n / 3 \rfloor} \lor \overline{K_{\lceil 2n / 3 \rceil}}$ attains the maximum Mostar index among all connected simple graphs of order $n$.
\end{conjecture}

Since Conjecture \ref{mostar_conj} involves an extremal problem, the choice of graph invariant function is clear: $G \mapsto \mathrm{Mo}(G)$. The following code snippet shows how this function can quickly be implemented using \texttt{SageMath}.

\begin{lstlisting}[language = Python, frame = trBL, escapeinside={(*@}{@*)}, aboveskip=10pt, belowskip=10pt, numbers=left, rulecolor=\color{black}]
def mostar_index(graph_batch: rlgt_graphs.Graph) -> np.ndarray:
    scores = np.empty(graph_batch.batch_size, dtype=np.float32)

    for index in range(graph_batch.batch_size):
        g = Graph(matrix(graph_batch.adjacency_matrix_colors[index]))
        if not g.is_connected():
            scores[index] = -2000.0
            continue

        transmissions = [sum(row) for row in g.distance_matrix().rows()]

        mostar = 0
        for u, v, _ in g.edges():
            mostar += abs(transmissions[u] - transmissions[v])

        scores[index] = mostar

    return scores
\end{lstlisting}

\begin{figure}[H]
\centering
\includegraphics[width=0.99\textwidth]{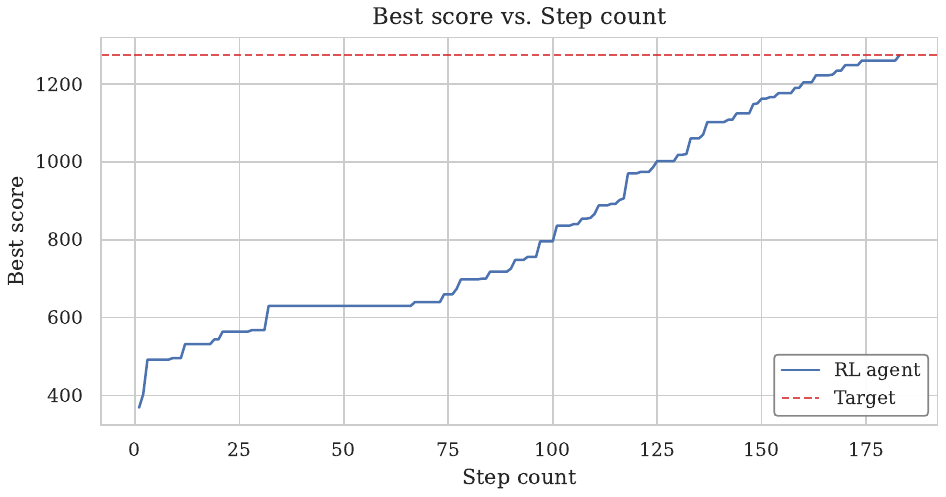}
\caption{The best score versus step count plot for the unsuccessful attempt to disprove Conjecture~\ref{mostar_conj} for the case $n = 21$.}
\label{plot_3}
\end{figure}

By configuring the Deep Cross-Entropy agent in conjunction with the Linear Build environment in largely the same manner as in Subsections \ref{appl_1} and \ref{appl_2}, we could not disprove Conjecture \ref{mostar_conj}. However, the agent did show clear signs of learning, and for many small values of $n \in \mathbb{N}$, say $n \le 24$, by restarting the learning process sufficiently many times, we could reach the greatest achieved graph invariant value of exactly $\mathrm{Mo}\left( K_{\lfloor n / 3 \rfloor} \lor \overline{K_{\lceil 2n / 3 \rceil}} \right)$ or close to this number. For example, the best score versus step count plot for the case $n = 21$ is shown in Figure~\ref{plot_3}; we do not provide the obtained graph because it is merely isomorphic to $K_7 \lor \overline{K_{14}}$. Since Conjecture~\ref{mostar_conj} can quickly be verified for each $n \in \{ 3, 4, 5, \ldots, 11 \}$ by using the \texttt{geng} tool from the package \texttt{nauty} \cite{McKayPip2014} together with \texttt{SageMath}, this provides further evidence that the conjecture may be true. Therefore, even if the RLGT framework cannot refute a conjecture of interest, it could still provide insight into whether the conjecture holds. The full source code for this \texttt{SageMath} script that attempted to disprove Conjecture \ref{mostar_conj} is available in the file \texttt{mostar\_index\_attempter.py}.

\section{Conclusion}\label{sc_conclusion}

In addition to being computationally efficient, the main advantage of the presented RL framework over existing approaches lies in its modularity, expressive power and ease of use. The notion of a graph is treated independently of the RL environment logic, and automatic conversions between several natural graph formats are supported. The agent--environment interaction logic is fully encapsulated on the agent side, with the agent implemented as a separate entity rather than being interwoven with the environment logic, as is common in many existing frameworks. Moreover, the RLGT framework supports directed graphs and graphs with more than two edge colors, which were not considered in earlier approaches.

The framework includes nine different RL environments implemented as seven classes inheriting from \texttt{GraphEnvironment}, and three RL methods implemented as classes inheriting from \texttt{GraphAgent}. The computational results from Section \ref{sc_applications} show that the Deep Cross-Entropy agent combined with the Linear Build environment performs particularly well, with comparable results obtainable using the other linear environments. While other agent--environment combinations can yield positive results, as shown in Subsection \ref{appl_1}, the learning process is generally less stable. For example, when disproving Conjecture \ref{agent_conj} using a global or local environment, satisfactory performance requires the use of the REINFORCE or PPO agent together with an initial graph generator that produces cycle graphs. Without these adjustments, the non-linear environments do not provide reliable results. Similarly, the REINFORCE and PPO agents do not perform well with the linear environments. This behavior may partly be due to the nature of the graph invariant functions considered in Section \ref{sc_applications}, all of which assign a large negative value to disconnected graphs. Such functions could destabilize training procedures that rely on discounted returns.

These observations suggest several directions for future research. One natural question is how the RL agents for extremal graph theory problems can be further refined. In particular, it could be useful to investigate modifications of the REINFORCE or PPO methods that improve their stability and performance in this setting. More broadly, exploring alternative RL methods for extremal graph theory applications remains an open and promising avenue. Since the choice of agent depends strongly on the specific problem, the modular design of the proposed framework facilitates the implementation of new approaches that can be used in conjunction with the existing RL environments.

Another natural direction for future research concerns the development of new RL environments for extremal graph theory. As discussed by Angileri et~al.\ \cite{Angileri2025}, the choice of environment can significantly influence the optimization process. Designing alternative environments, including those with nondeterministic RL tasks, may therefore prove valuable. The proposed framework does not impose deterministic assumptions on environments, and can thus be extended to accommodate such settings.

\section*{Acknowledgments}

The authors are grateful to Nino Bašić for his useful comments and suggestions.

\section*{Conflict of interest}

The authors declare that they have no conflict of interest.


\begin{thebibliography}{99}

\bibitem{TensorFlow}
M.\ Abadi, A.\ Agarwal, P.\ Barham, E.\ Brevdo, Z.\ Chen, C.\ Citro, G.\ S.\ Corrado, A.\ Davis, J.\ Dean, M.\ Devin, S.\ Ghemawat, I.\ Goodfellow, A.\ Harp, G.\ Irving, M.\ Isard, Y.\ Jia, R.\ Józefowicz, Ł.\ Kaiser, M.\ Kudlur, J.\ Levenberg, D.\ Mané, R.\ Monga, S.\ Moore, D.\ Murray, C.\ Olah, M.\ Schuster, J.\ Shlens, B.\ Steiner, I.\ Sutskever, K.\ Talwar, P.\ Tucker, V.\ Vanhoucke, V.\ Vasudevan, F.\ Viégas, O.\ Vinyals, P.\ Warden, M.\ Wattenberg, M.\ Wicke, Y.\ Yu and X.\ Zheng, TensorFlow: Large-scale machine learning on heterogeneous distributed systems, 2016, \arxiv{1603.04467}{cs.DC}.

\bibitem{AkAlaAn2021}
S.\ Akbari, A.\ Alazemi and M.\ Anđelić, Upper bounds on the energy of graphs in terms of matching number, {\em Appl.\ Anal.\ Discrete Math.\/}\ {\bf 15} (2021), 444--459, \doi{10.2298/AADM201227016A}.

\bibitem{AliDo2021}
A.\ Ali and T.\ Došlić, Mostar index: Results and perspectives, {\em Appl.\ Math.\ Comput.\/}\ {\bf 404} (2021), 126245, \doi{10.1016/j.amc.2021.126245}.

\bibitem{Angileri2025}
F.\ Angileri, G.\ Lombardi, A.\ Fois, R.\ Faraone, C.\ Metta, M.\ Salvi, L.\ A.\ Bianchi, M.\ Fantozzi, S.\ G.\ Galfrè, D.\ Pavesi, M.\ Parton and F.\ Morandin, A systematization of the Wagner framework: Graph theory conjectures and reinforcement learning, in: D.\ Pedreschi, A.\ Monreale, R.\ Guidotti, R.\ Pellungrini and F.\ Naretto (eds.), {\em Discovery Science}, volume 15243 of {\em Lecture Notes in Computer Science}, Springer, Cham, 2025, pp.\ 325--338, \doi{10.1007/978-3-031-78977-9_21}.

\bibitem{Angileri2025B}
F.\ Angileri, G.\ Lombardi, A.\ Fois, R.\ Faraone, C.\ Metta, M.\ Salvi, L.\ A.\ Bianchi, M.\ Fantozzi, S.\ G.\ Galfrè, D.\ Pavesi, M.\ Parton and F.\ Morandin, Analyzing RL components for Wagner's framework via Brouwer's conjecture, {\em Mach.\ Learn.\/}\ {\bf 114} (2025), Art.\ No.\ 242, \doi{10.1007/s10994-025-06890-2}.

\bibitem{Biggs1993}
N.\ Biggs, {\em Algebraic Graph Theory}, 2nd edition, Cambridge Mathematical Library, Cambridge University Press, Cambridge, 1993, \doi{10.1017/CBO9780511608704}.

\bibitem{BoKroMaRu2005}
P.-T.\ de Boer, D.\ P.\ Kroese, S.\ Mannor and R.\ Y.\ Rubinstein, A tutorial on the cross-entropy method, {\em Ann.\ Oper.\ Res.\/}\ {\bf 134} (2005), 19--67, \doi{10.1007/s10479-005-5724-z}.

\bibitem{Bollobas1998}
B.\ Bollobás, \emph{Modern Graph Theory}, volume 184 of \emph{Graduate Texts in Mathematics}, Springer, New York, NY, 1998, \doi{10.1007/978-1-4612-0619-4}.

\bibitem{BonMur1976}
J.\ A.\ Bondy and U.\ S.\ R.\ Murty, {\em Graph Theory with Applications}, Elsevier Science Publishing Co., Inc., New York, NY, 1976.

\bibitem{BraHaSte2006}
V.\ Brankov, P.\ Hansen and D.\ Stevanović, Automated conjectures on upper bounds for the largest Laplacian eigenvalue of graphs, {\em Linear Algebra Appl.\/}\ {\bf 414} (2006), 407--424, \doi{10.1016/j.laa.2005.10.017}.

\bibitem{BrouHae2012}
A.\ E.\ Brouwer and W.\ H.\ Haemers, {\em Spectra of Graphs}, Universitext, Springer, New York, NY, 2012, \doi{10.1007/978-1-4614-1939-6}.

\bibitem{CaoChenJingStieToft2019}
Y.\ Cao, G.\ Chen, G.\ Jing, M.\ Stiebitz and B.\ Toft, Graph edge coloring: A survey, {\em Graphs Combin.\/}\ {\bf 35} (2019), 33--66, \doi{10.1007/s00373-018-1986-5}.

\bibitem{Isort}
T.\ Crosley, et al., \texttt{isort}, \url{https://pycqa.github.io/isort/}.

\bibitem{CvetDoobSachs1995}
D.\ Cvetković, M.\ Doob and H.\ Sachs, \emph{Spectra of Graphs: Theory and Application}, 2nd edition, Johann Ambrosius Barth Verlag, Heidelberg--Leipzig, 1980.

\bibitem{Documentation}
I.\ Damnjanović, U.\ Milivojević, I.\ Đorđević and D.\ Stevanović, RLGT: A reinforcement learning framework for extremal graph theory (GitHub documentation), \url{https://ivan-damnjanovic.github.io/rlgt/}.

\bibitem{GitHub}
I.\ Damnjanović, U.\ Milivojević, I.\ Đorđević and D.\ Stevanović, RLGT: A reinforcement learning framework for extremal graph theory (GitHub repository), \url{https://github.com/Ivan-Damnjanovic/rlgt}.

\bibitem{PyPI}
I.\ Damnjanović, U.\ Milivojević, I.\ Đorđević and D.\ Stevanović, RLGT: A reinforcement learning framework for extremal graph theory (PyPI page), \url{https://pypi.org/project/RLGT/}.

\bibitem{Diestel2017}
R.\ Diestel, \emph{Graph Theory}, 5th edition, volume 173 of \emph{Graduate Texts in Mathematics}, Springer Berlin, Heidelberg, 2017, \doi{10.1007/978-3-662-53622-3}.

\bibitem{DoMaSkrTiZu2018}
T.\ Došlić, I.\ Martinjak, R.\ Škrekovski, S.\ Tipurić Spužević and I.\ Zubac, Mostar index, {\em J.\ Math.\ Chem.\/}\ {\bf 56} (2018), 2995--3013, \doi{10.1007/s10910-018-0928-z}.

\bibitem{Erdos1975}
P.\ Erdős, Some recent progress on extremal problems in graph theory, {\em Congr.\ Numer.\/}\ {\bf 14} (1975), 3--14. 

\bibitem{Poetry}
S.\ Eustace, et al., \texttt{Poetry}, \url{https://python-poetry.org/}.

\bibitem{GheYaKaSte2026}
M.\ Ghebleh, S.\ Al-Yakoob, A.\ Kanso and D.\ Stevanović, Graphs having two main eigenvalues and arbitrarily many distinct vertex degrees, {\em Appl.\ Math.\ Comput.\/}\ {\bf 495} (2025), 129311, \doi{10.1016/j.amc.2025.129311}.

\bibitem{GheYaKaSte2025}
M.\ Ghebleh, S.\ Al-Yakoob, A.\ Kanso and D.\ Stevanović, Reinforcement learning for graph theory, II.\ Small Ramsey numbers, {\em Art Discrete Appl.\ Math.\/}\ {\bf 8} (2025), \#P1.07, \doi{10.26493/2590-9770.1788.8af}.

\bibitem{GheYaKaSte2024}
M.\ Ghebleh, S.\ Al-Yakoob, A.\ Kanso and D.\ Stevanović, Reinforcement learning for graph theory, I: Reimplementation of Wagner's approach, {\em Discrete Appl.\ Math.\/}\ {\bf 380} (2026), 468--479, \doi{10.1016/j.dam.2025.10.047}.

\bibitem{Graph6Java}
M.\ Ghebleh, A.\ Kanso and D.\ Stevanović, Graph6Java: A researcher-friendly Java framework for testing conjectures in chemical graph theory, {\em MATCH Commun.\ Math.\ Comput.\ Chem.\/}\ {\bf 81} (2019) 737--770, \url{https://match.pmf.kg.ac.rs/electronic_versions/Match81/n3/match81n3_737-770.pdf}.

\bibitem{Glover1989}
F.\ Glover, Tabu search --- Part I, {\em ORSA J.\ Comput.\/}\ {\bf 1} (1989), 190--206, \doi{10.1287/ijoc.1.3.190}.

\bibitem{GloLa1998}
F.\ Glover and M.\ Laguna, Tabu search, in: D.-Z.\ Du and P.\ M.\ Pardalos (eds.), {\em Handbook of Combinatorial Optimization}, Springer, Boston, MA, 1998, pp.\ 2093--2229, \doi{10.1007/978-1-4613-0303-9_33}.

\bibitem{GodRoy2001}
C.\ Godsil and G.\ Royle, {\em Algebraic Graph Theory}, volume 207 of {\em Graduate Texts in Mathematics}, Springer, New York, NY, 2001, \doi{10.1007/978-1-4613-0163-9}.

\bibitem{Gutman1978}
I.\ Gutman, The energy of a graph, {\em Ber.\ Math.-Statist.\ Sekt.\ Forschungsz.\ Graz.\/}\ {\bf 103} (1978), 1--22.

\bibitem{NumPy}
C.\ R.\ Harris, K.\ J.\ Millman, S.\ J.\ van der Walt, R.\ Gommers, P.\ Virtanen, D.\ Cournapeau, E.\ Wieser, J.\ Taylor, S.\ Berg, N.\ J.\ Smith, R.\ Kern, M.\ Picus, S.\ Hoyer, M.\ H.\ van Kerkwijk, M.\ Brett, A.\ Haldane, J.\ Fernández del Río, M.\ Wiebe, P.\ Peterson, P.\ Gérard-Marchant, K.\ Sheppard, T.\ Reddy, W.\ Weckesser, H.\ Abbasi, C.\ Gohlke and T.\ E.\ Oliphant, Array programming with NumPy, {\em Nature} {\bf 585} (2020), 357--362, \doi{10.1038/s41586-020-2649-2}.

\bibitem{Pytest}
H.\ Krekel, et al., \texttt{pytest}, \url{https://docs.pytest.org/}.

\bibitem{Black}
Ł.\ Langa, et al., \texttt{Black}, \url{https://black.readthedocs.io/en/stable/}.

\bibitem{Lapan2020}
M.\ Lapan, \emph{Deep Reinforcement Learning Hands-On}, 2nd edition, Packt Publishing, Birmingham, 2020.

\bibitem{MaSviIvBu2021}
N.\ Mazyavkina, S.\ Sviridov, S.\ Ivanov and E.\ Burnaev, Reinforcement learning for combinatorial optimization: A survey, \emph{Comput.\ Oper.\ Res.\/}\ {\bf 134} (2021), 105400, \doi{10.1016/j.cor.2021.105400}.

\bibitem{McKayPip2014}
B.\ D.\ McKay and A.\ Piperno, Practical graph isomorphism, II, {\em J.\ Symb.\ Comput.\/}\ {\bf 60} (2014) 94--112, \doi{10.1016/j.jsc.2013.09.003}.

\bibitem{Mehrabian2024}
A.\ Mehrabian, A.\ Anand, H.\ Kim, N.\ Sonnerat, M.\ Balog, Gh.\ Comanici, T.\ Berariu, A.\ Lee, A.\ Ruoss, A.\ Bulanova, D.\ Toyama, S.\ Blackwell, B.\ Romera Paredes, P.\ Veličković, L.\ Orseau, J.\ Lee, A.\ M.\ Naredla, D.\ Precup and A.\ Zs.\ Wagner, Finding increasingly large extremal graphs with AlphaZero and tabu search, in: K.\ Larson (ed.), {\em Proceedings of the Thirty-Third International Joint Conference on Artificial Intelligence}, IJCAI Organization, 2024, pp.\ 6985--6993, \doi{10.24963/ijcai.2024/772}.

\bibitem{JPype}
S.\ Menard, L.\ Nell, et al., \texttt{JPype}, \url{https://jpype.readthedocs.io/en/latest/}.

\bibitem{NanzFuria2015}
S.\ Nanz and C.\ A.\ Furia, A comparative study of programming languages in Rosetta Code, 2015, \arxiv{1409.0252}{cs.SE}.

\bibitem{PyTorch}
A.\ Paszke, S.\ Gross, F.\ Massa, A.\ Lerer, J.\ Bradbury, G.\ Chanan, T.\ Killeen, Z.\ Lin, N.\ Gimelshein, L.\ Antiga, A.\ Desmaison, A.\ Köpf, E.\ Yang, Z.\ DeVito, M.\ Raison, A.\ Tejani, S.\ Chilamkurthy, B.\ Steiner, L.\ Fang, J.\ Bai and S.\ Chintala, PyTorch: An imperative style, high-performance deep learning library, 2019, \arxiv{1912.01703}{cs.LG}.

\bibitem{Powell2022}
W.\ B.\ Powell, \emph{Reinforcement Learning and Stochastic Optimization: A Unified Framework for Sequential Decisions}, 1st edition, John Wiley \& Sons, Inc., Hoboken, NJ, 2022, \doi{10.1002/9781119815068}.

\bibitem{Radziszowski}
S.\ P.\ Radziszowski, Small Ramsey numbers, {\em Electron.\ J.\ Combin.\/}\ {\bf DS01} (2024), \doi{10.37236/21}.

\bibitem{Rowlinson2007}
P.\ Rowlinson, The main eigenvalues of a graph: A survey, {\em Appl.\ Anal.\ Discrete Math.\/}\ {\bf 1} (2007), 445--471, \doi{10.2298/AADM0702445R}.

\bibitem{Rubinstein1997}
R.\ Y.\ Rubinstein, Optimization of computer simulation models with rare events, {\em Eur.\ J.\ Oper.\ Res.\/}\ {\bf 99} (1997), 89--112, \doi{10.1016/S0377-2217(96)00385-2}.

\bibitem{Schulman2017}
J.\ Schulman, F.\ Wolski, P.\ Dhariwal, A.\ Radford and O.\ Klimov, Proximal Policy Optimization algorithms, 2017, \arxiv{1707.06347}{cs.LG}.

\bibitem{Silver2017}
D.\ Silver, J.\ Schrittwieser, K.\ Simonyan, I.\ Antonoglou, A.\ Huang, A.\ Guez, T.\ Hubert, L.\ Baker, M.\ Lai, A.\ Bolton, Y.\ Chen, T.\ Lillicrap, F.\ Hui, L.\ Sifre,
G.\ van den Driessche, T.\ Graepel and D.\ Hassabis, Mastering the game of Go without human knowledge, {\em Nature} {\bf 550} (2017), 354--359, \doi{10.1038/nature24270}.

\bibitem{Simonovits1984}
M.\ Simonovits, Extremal graph problems, degenerate extremal problems, and supersaturated graphs, in: J.\ A.\ Bondy and U.\ S.\ R.\ Murty (eds.), {\em Progress in Graph Theory},
Academic Press, Toronto, ON, 1984, pp.\ 419--437.

\bibitem{ShaRe2020}
R.\ Sharafdini and T.\ Réti, On the transmission-based graph topological indices, {\em Kragujevac J.\ Math.\/}\ {\bf 44} (2020), 41--63, \doi{10.46793/KgJMat2001.041S}.

\bibitem{SoIoRoSe2022}
P.\ Soviany, R.\ T.\ Ionescu, P.\ Rota and N.\ Sebe, Curriculum learning: A survey, {\em Int.\ J.\ Comput.\ Vision} {\bf 130} (2022), 1526--1565, \doi{10.1007/s11263-022-01611-x}.

\bibitem{SteDaSte2021}
Đ.\ Stevanović, I.\ Damnjanović and D.\ Stevanović, Finding counterexamples for a conjecture of Akbari, Alazemi and Anđelić, 2021, \arxiv{2111.15303}{math.CO}.

\bibitem{SuBa2018}
R.\ S.\ Sutton and A.\ G.\ Barto, \emph{Reinforcement Learning: An Introduction}, 2nd edition, Adaptive Computation and Machine Learning, MIT Press, Cambridge, MA, 2018.

\bibitem{Szepesvari2010}
Cs.\ Szepesvári, \emph{Algorithms for Reinforcement Learning}, 1st edition, Synthesis Lectures on Artificial Intelligence and Machine Learning, Springer, Cham, 2010, \doi{10.1007/978-3-031-01551-9}.

\bibitem{TaRouCaHa2025}
L.\ Taieb, M.\ Roucairol, T.\ Cazenave and A.\ Harutyunyan, Automated refutation with Monte Carlo search of graph theory conjectures on the maximum Laplacian eigenvalue, in: Y. Zhang, M. Hladík and H. Moosaei (eds.), \emph{Learning and Intelligent Optimization}, volume 15744 of \emph{Lecture Notes in Computer Science}, Springer, Cham, 2026, pp.\ 52--63, \doi{10.1007/978-3-032-09156-7_4}.

\bibitem{Gymnasium}
M.\ Towers, A.\ Kwiatkowski, J.\ Terry, J.\ U.\ Balis, G.\ De Cola, T.\ Deleu, M.\ Goulão, A.\ Kallinteris, M.\ Krimmel, A.\ KG, R.\ Perez-Vicente, A.\ Pierré, S.\ Schulhoff, J.\ J.\ Tai, H.\ Tan and O.\ G.\ Younis, Gymnasium: A standard interface for reinforcement learning environments, 2025, \arxiv{2407.17032}{cs.LG}.

\bibitem{Wagner2021}
A.\ Zs.\ Wagner, Constructions in combinatorics via neural networks, 2021, \arxiv{2104.14516}{math.CO}.

\bibitem{Williams1992}
R.\ J.\ Williams, Simple statistical gradient-following algorithms for connectionist reinforcement learning, {\em Mach.\ Learn.\/}\ {\bf 8} (1992), 229--256, \doi{10.1007/BF00992696}.

\bibitem{SageMath}
The Sage Developers, SageMath, the Sage Mathematics Software System (Version 10.8), 2025, \url{https://www.sagemath.org}.

\end{thebibliography}
\end{document}